\documentclass[preprint,12pt]{elsarticle}

\usepackage{amssymb}
\usepackage{amsmath}
\usepackage{times}
\usepackage{soul}
\usepackage[table]{xcolor}
\usepackage{url}
\usepackage[utf8]{inputenc}
\usepackage[small]{caption}
\usepackage{graphicx}
\usepackage{amsthm}
\usepackage{booktabs}
\usepackage{algorithm}
\usepackage{algorithmic}
\usepackage[switch]{lineno}
\usepackage{amssymb}
\usepackage{subcaption}
\usepackage{enumitem}
\usepackage{multirow}
\usepackage{subfiles} 
\usepackage{pdfpages}

\newtheorem{assumption}{Assumption}

\journal{Knowledge-Based Systems}

\begin{document}

\begin{frontmatter}

%% Title, authors and addresses

%% use the tnoteref command within \title for footnotes;
%% use the tnotetext command for theassociated footnote;
%% use the fnref command within \author or \affiliation for footnotes;
%% use the fntext command for theassociated footnote;
%% use the corref command within \author for corresponding author footnotes;
%% use the cortext command for theassociated footnote;
%% use the ead command for the email address,
%% and the form \ead[url] for the home page:
%% \title{Title\tnoteref{label1}}
%% \tnotetext[label1]{}
%% \author{Name\corref{cor1}\fnref{label2}}
%% \ead{email address}
%% \ead[url]{home page}
%% \fntext[label2]{}
%% \cortext[cor1]{}
%% \affiliation{organization={},
%%             addressline={},
%%             city={},
%%             postcode={},
%%             state={},
%%             country={}}
%% \fntext[label3]{}

\title{AFBS:Buffer Gradient Selection in Semi-asynchronous Federated Learning} %% Article title

%% use optional labels to link authors explicitly to addresses:
%% \author[label1,label2]{}
%% \affiliation[label1]{organization={},
%%             addressline={},
%%             city={},
%%             postcode={},
%%             state={},
%%             country={}}
%%
%% \affiliation[label2]{organization={},
%%             addressline={},
%%             city={},
%%             postcode={},
%%             state={},
%%             country={}}
% \author[1]{Yiting Zheng}
\author[1,2]{Chaoyi Lu}\ead{chaoyi@stu.xjtu.edu.cn} %% Author name
\author[1]{Yiding Sun}\ead{sunyiding@stu.xjtu.edu.cn} %% Author name
\author[1]{Jinqian Chen}\ead{chenjinqian@stu.xjtu.edu.cn} %% Author name
\author[1]{Zhichuan Yang}\ead{zhichuan@stu.xjtu.edu.cn}
\author[3]{Jiangming Pan}\ead{panjiangming@kuaishou.com}
\author[1,2]{Jihua Zhu\cormark[cor1]}\ead{zhujh@xjtu.edu.cn}
\cortext[cor1]{Corresponding Author}
%% Author affiliation
\affiliation[1]{organization={School of Software Engineering},
                addressline={Xi'an Jiaotong University}, 
                city={Xi'an},
%               citysep={}, % Uncomment if no comma needed between city and postcode
                postcode={710049}, 
                country={China}}
\affiliation[2]{organization={State Key Laboratory of Integrated Services Networks},
                addressline={Xidian University}, 
                city={Xi'an},
%               citysep={}, % Uncomment if no comma needed between city and postcode
                postcode={710071}, 
                country={China}}
\affiliation[3]{
                addressline={Kuaishou Technology}, 
                city={Beijing},
%               citysep={}, % Uncomment if no comma needed between city and postcode
                postcode={100085}, 
                country={China}}

%% Abstract
\begin{abstract}
Asynchronous federated learning (AFL) accelerates training by eliminating the need to wait for stragglers. However, its asynchronous nature introduces gradient staleness, where outdated gradients degrade performance. Existing solutions address this issue with gradient buffers, forming a semi-asynchronous framework. Nevertheless, this approach struggles when buffers accumulate numerous stale gradients, as blindly aggregating all gradients can harm training. To address this, we propose \textit{AFBS} (\underline{A}synchronous \underline{F}L \underline{B}uffer \underline{S}election), the first algorithm to perform gradient selection within buffers while ensuring privacy protection. Specifically, the client sends the random projection encrypted label distribution matrix before training, and the server performs client clustering based on it. During training, server scores and selects gradients within each cluster based on their informational value, discarding low-value gradients to enhance semi-asynchronous federated learning. Extensive experiments in highly heterogeneous system and data environments demonstrate AFBS's superior performance compared to state-of-the-art methods. Notably, on the most challenging task, CIFAR-100, AFBS improves accuracy by up to 4.8\% over the previous best algorithm and reduces the time to reach target accuracy by 75\%.
\end{abstract}

%%Graphical abstract
\begin{graphicalabstract}
\begin{figure*}[h]
    \centering
    \begin{minipage}{1.0\textwidth}
        \includegraphics[width=\linewidth]{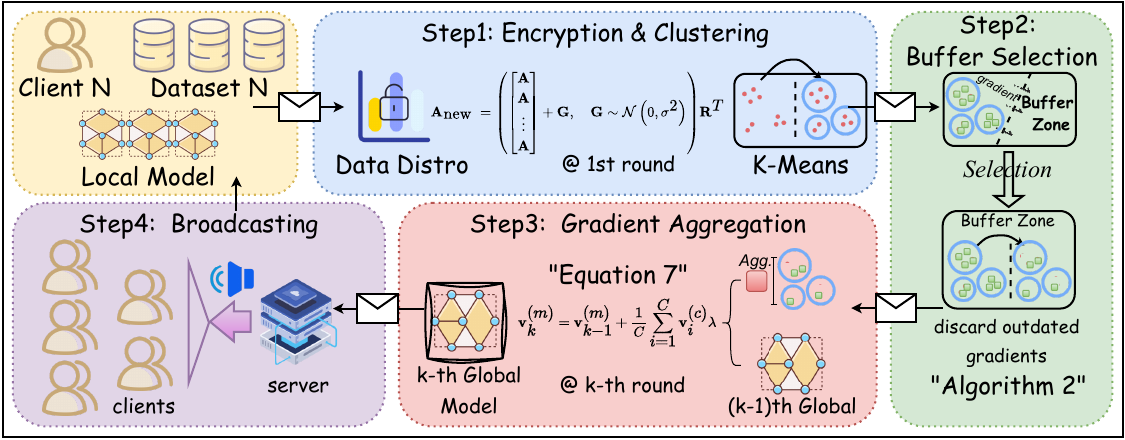}
    \end{minipage}%
\end{figure*}
    
\textbf{The AFBS framework.} Clients first need to encrypt the label distribution using random projection before training. The server will cluster the clients based on this information. During training, the server collects gradients sent by clients until the buffer is full. Once the buffer is full, the server employs the AFBS algorithm to perform Gradient Selection on the gradients within each cluster. The remaining gradients in the buffer are then aggregated and combined once more with the global model. The server broadcasts this new global model to the clients, commencing a new round of federated learning.

\end{graphicalabstract}

%%Research highlights
\begin{highlights}
\item We propose AFBS, the first algorithm to perform gradient selection on semi-asynchronous federated learning buffers. By eliminating low-value stale gradients from the buffer, it significantly improves final accuracy and training speed. Moreover, the reduced number of gradients participating in aggregation decreases costly gradient summation operations, substantially lowering computational costs on the server side.
\item We propose a novel encryption algorithm. Initially, the clients' data distribution is transformed into a full-rank matrix through the addition of Gaussian noise, followed by information encryption via random projection. This method is also applicable to other scenarios requiring matrix encryption.
\item We employ rigorous mathematical theories to prove the convergence of the proposed method and the soundness of the encryption approach. Furthermore, extensive experiments are conducted to demonstrate the superiority of the proposed method.
\end{highlights}

%% Keywords
\begin{keyword}
Federated Learning \sep
Asynchronous Aggregation \sep
Gradient Selection
\end{keyword}

\end{frontmatter}

\section{Introduction}

Federated learning, as a machine learning paradigm with privacy protection, is becoming increasingly popular \cite{Yang2019,10.1007/978-3-030-58607-2_5}. It transforms data transmission into the transmission of gradients or models, thus avoiding privacy issues, and it improves accuracy compared to local training. Federated learning is orchestrated by a server that coordinates the clients to participate in the training process. At each round, a subset of clients is selected, and the global model is broadcast to them. The selected clients then train the model using their local data. After training is completed on each selected client, the server aggregates all the parameters uploaded by the clients to generate a new global model. In traditional synchronous federated learning, every round requires waiting for all clients to finish training, which can lead to significant waiting issues. This is especially problematic when there is a large performance disparity between clients, as the fastest client often ends up waiting for the slowest client, with the waiting time being much longer than the training time of the fastest client. This issue is referred to as the ``straggler" problem \cite{9887795,ijcai2024p419,lu2025correctedlatestversionmake}.

\begin{figure}[t]
    \centering
    \includegraphics[width=1\textwidth]{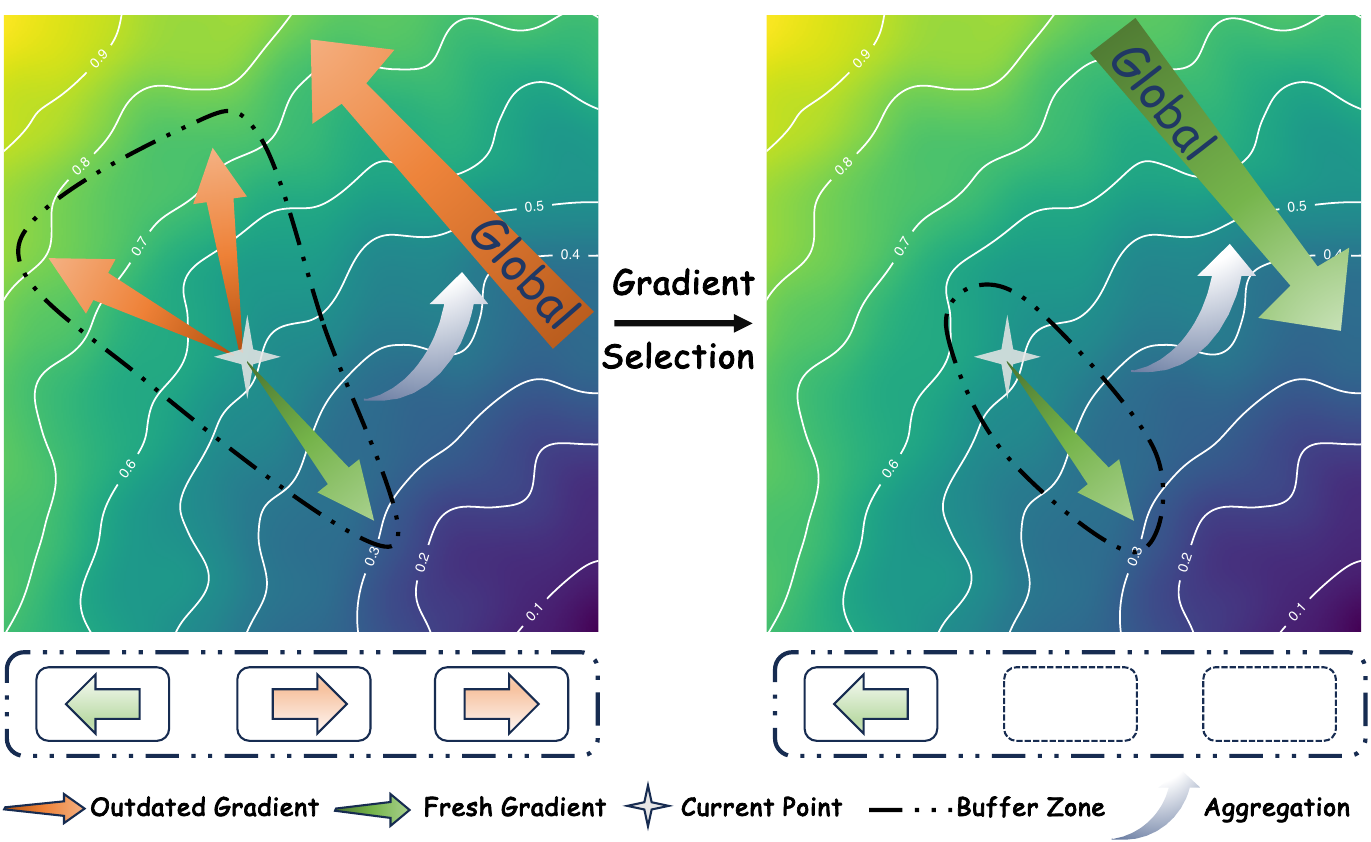}  % 导入 PDF 文件

    \caption{Schematic diagram of buffer gradient selection. According to the gradient selection strategy, outdated gradients are removed from the buffer, thereby redirecting the gradient direction towards the optimal solution.}  
    \vspace{-0.5cm}
    \label{fig:figure1}
\end{figure}

Several factors may cause the straggler problem, with the most significant reasons being system heterogeneity and communication heterogeneity \cite{kairouz2021advancesopenproblemsfederated}. AFL alleviates the straggler problem: the asynchronous approach no longer waits for stragglers and aggregates gradients as soon as they are uploaded by the clients. However, due to issues like staleness and data heterogeneity~\cite{CHEN20211}, the accuracy of asynchronous methods is often lower. While synchronous federated learning ensures uniform client versions across rounds, AFL concurrently aggregates models from divergent versions. Such version mixing carries risks: When stale models combine with the current global model, training integrity may be compromised. These dynamics ultimately introduce heightened diversity and complexity into AFL's update process.

Given the considerable negative consequences associated with obsolescence, is it feasible to discard outdated gradients and exclusively select those that are advantageous? This question has drawn our attention to the process of client selection in federated learning~\cite{fu2023clientselectionfederatedlearning,9212434}. This field has recognized that not all clients contribute positively to the training process. Clients with poor performance, small datasets, or slow communication often drag down the entire training process. Therefore, it is necessary to select high-performing clients for training and exclude those with suboptimal performance. Similarly, the gradients in the buffer are comparable to the client selection problem. It is not always optimal to aggregate all the gradients in the buffer, as some may be detrimental to the training, such as severely outdated gradients. When comparing two gradients with different levels of staleness, if they come from clients with similar data distributions, the older gradient becomes unnecessary, as the newer gradient better represents the correct gradient direction for the data distribution in recent rounds. However, as a privacy-preserving framework, federated learning typically faces the issue that clients refuse to directly provide their data distribution to the server, as this could lead to privacy leakage~\cite{9599369}. Therefore, how to represent the clients' data distribution in alternative ways and accurately determine the similarity between clients in an asynchronous environment remains a challenge.

Although staleness has a significant negative impact on AFL, most existing work mitigates staleness by applying decay weights to the magnitude of gradient updates ~\cite{xie2020asynchronousfederatedoptimization,8945292}. However, these approaches do not focus on buffer management algorithms for semi-asynchronous federated learning, which makes it difficult for semi-asynchronous algorithms to fully utilize their advantages. For a gradient with severe staleness, even though the decay weight algorithm reduces its magnitude significantly, it still participates in aggregation. Since all gradients are averaged in the end, this outdated gradient still reduces the weight of other normal gradients, thereby decreasing the step size of each update and limiting the performance of semi-asynchronous federated learning. To address this issue, we propose a specialized algorithm to optimize semi-asynchronous federated learning by no longer retaining gradients with clearly low value in the buffer. This accelerates the convergence of semi-asynchronous federated learning and improves its accuracy. The gradient selection concept is shown in Figure \ref{fig:figure1}.

In this paper, we propose a gradient scoring mechanism and a gradient selection algorithm, allowing semi-asynchronous federated learning to aggregate more valuable gradients, improving both accuracy and convergence speed. The main contributions of this paper are as follows:
\begin{itemize}
    \item We propose a simple and efficient algorithm AFBS, which integrates gradient scoring mechanism and gradient selection, and significantly improves the accuracy and convergence rate of semi asynchronous federated learning by utilizing the unique advantages of buffer. Furthermore, it significantly reduces the computational cost on the server side.
    \item We investigate the impact of client similarity determination in asynchronous scenarios and propose a privacy-preserving client clustering strategy based on random projection.
    \item We compare AFBS with six other methods on four representative datasets, and experimental results demonstrate that AFBS exhibits highly competitive performance, and demonstrate significant advantages in highly challenging tasks.
\end{itemize}

\section{Related Work}

\subsection{Synchronous Federated Learning}
Synchronous methods are the primary approach in federated learning, characterized by the server sampling a subset of clients to participate in training during each communication round. The server only aggregates the uploaded model information after all sampled clients complete their training. Synchronous methods have significant advantages in terms of model accuracy. Among them, FedAvg~\cite{pmlr-v54-mcmahan17a} is the most classical algorithm in synchronous federated learning, which generates a new global model in each communication round by averaging the models from all sampled clients. FedAdam 
 ~\cite{reddi2021adaptivefederatedoptimization}, on the other hand, introduces a momentum mechanism based on FedAvg to further enhance optimization performance.

Another notable advantage of synchronous federated learning is that clients always train based on the latest global model, while the server receives the most up-to-date gradient information. Consequently, synchronous algorithms can effectively collect the latest client data and optimize the global model to address various complex challenges. For instance, FedProx \cite{li2020federatedoptimizationheterogeneousnetworks} adds a regularization term during local client training to prevent model parameters from deviating excessively from the global model. FedBN ~\cite{li2021fedbnfederatedlearningnoniid} alleviates feature shift caused by data distribution differences through Batch Normalization. Meanwhile, MOON~\cite{9578660} leverages the similarity between model representations to correct the local training of individual clients. In contrast, semi-asynchronous federated learning aggregates gradients from different versions each time, posing significant challenges to the training process.

\begin{table}[t]
\centering
\caption{Frequently used notations in this article.}
\label{tab:notations}
\begin{tabular}{cl}
\toprule
\textbf{Symbol} & \textbf{Description} \\ 
\midrule
$K$            & number of client clusters \\
$S$          & the buffer\\ 
$C$        & the capacity of buffer\\ 
$Q$        & local update steps \\ 
$\tau$    & staleness \\ 
$\tau^{\text{min}}$ & the min staleness in the buffer\\ 
$\eta_g$       & global learning rate \\ 
$\mathcal{V}$ &the dataset size of the gradient source client\\
$T$            & communication rounds \\ 
$\alpha$       & alpha in the Dirichlet distribution \\ 
$\Delta_t^k$   & accumulated gradients at $t_k$ steps from client $k$ \\ 
$R$  &Random projection matrix\\
$G$ &Gaussian noise matrix\\
$\epsilon$ &the allowable error margin\\
$\lambda$ &obsolescence decay factor\\
\bottomrule
\end{tabular}
\end{table}

\subsection{Asynchronous Federated Learning}
AFL is primarily designed to address the straggler problem, which refers to the negative impact caused by fast clients waiting excessively for slow clients. In fully asynchronous methods, the global model is updated immediately upon receiving updates from clients, and clients are no longer required to wait for slower clients. This characteristic enables asynchronous methods to achieve significantly faster training speeds than synchronous methods, especially in scenarios with highly uneven client latency distributions. However, asynchronous methods often suffer from the issue of staleness, resulting in poor accuracy and stability, particularly under non-i.i.d. data distributions. To mitigate this issue, various strategies have been proposed. For instance, FedAsync \cite{xie2020asynchronousfederatedoptimization} assigns weights to updates based on their staleness, FedASMU \cite{Liu_Jia_Che_Huo_Ren_Zhou_Dai_Dou_2024} distributes the latest global model to clients during training, and CA2FL \cite{wang2024tackling} caches the most recent updates for each client to calibrate the global model.

In addition, several studies have proposed introducing a buffer at the server to collect and aggregate updates sent by clients, a method referred to as semi-asynchronous federated learning. For example, FedBuff \cite{nguyen2022federatedlearningbufferedasynchronous} clears the buffer once it is full, while FedFa \cite{10.24963/ijcai.2024/584} employs a queue-based buffer that removes the oldest update when a new update arrives after the buffer is full. Buffers can effectively alleviate the impact of staleness in AFL. However, harmful gradients or model updates that adversely affect training may still exist within the buffer. Removing such harmful updates prior to aggregation could further benefit the training process. Despite this potential improvement, existing works have not yet proposed algorithms to address this specific issue.
Our work makes the first attempt to explore this operation and experimentally demonstrates the advantages it brings.

\subsection{Cluster Federated Learning}
In real-world scenarios, the data distribution of clients can usually be divided into several distinct clusters, where the data distribution within each cluster is relatively similar, but there are significant differences in data distribution between clusters. By dividing clients into multiple clusters, Cluster FL enables federated learning tailored to the data distribution of each cluster, effectively mitigating the challenge of data heterogeneity in federated learning. Existing clustering algorithms in clustered federated learning can be broadly categorized into two types: one-time clustering and iterative clustering. One-time clustering completes the client division before training begins. For instance, \cite{ghosh2019robustfederatedlearningheterogeneous} employs a robust clustering method to divide all clients into K clusters, while PFA \cite{liu2021pfaprivacypreservingfederatedadaptation} utilizes privacy-preserving representations generated by neural networks for clustering. In contrast, iterative clustering dynamically updates clustering results during the training process. For example, the CFL method \cite{9174890} recursively separates two groups of clients with inconsistent descent directions to achieve clustering, whereas IFCA~\cite{ghosh2021efficientframeworkclusteredfederated} randomly generates cluster centers and assigns clients to clusters with the minimum loss value to complete clustering.

The proposed method draws on the idea of clustered federated learning, where clients are grouped based on data distribution before gradient selection. To ensure privacy preservation in federated learning, this paper also introduces a novel encryption approach, which effectively addresses the challenge of privacy protection in clustered federated learning.

\section{Method}
\subsection{Preliminary}
Consider a system comprising $M$ devices, represented by the set $\mathcal{N} = \{1, 2, \ldots, M\}$. Each device $i \in \mathcal{N}$ possesses a local dataset defined as:
\begin{equation}
S_i = \{(u_{i,k}, v_{i,k}) \mid k = 1, 2, \ldots, |S_i|\},
\end{equation}
where $u_{i,k} \in \mathbb{R}^s$ denotes the $k$-th input sample, $v_{i,k} \in \mathbb{R}$ is its corresponding label, and $|S_i|$ represents the number of samples on device $i$. Combining the datasets from all devices yields the global dataset:
\begin{equation}
S = \bigcup_{i \in \mathcal{N}} S_i, \quad T = \sum_{i \in \mathcal{N}} |S_i|,
\end{equation}
where $T$ is the total number of samples across all devices.

The objective is to collaboratively train a global model using the local datasets without directly sharing raw data. The global loss function is formulated as:
\begin{equation}
F(\theta) = \frac{1}{T} \sum_{i \in \mathcal{N}} \sum_{(u_{i,k}, v_{i,k}) \in S_i} \mathcal{L}(\theta; u_{i,k}, v_{i,k}),
\end{equation}
where $\mathcal{L}(\cdot)$ represents the loss for an individual sample. The local loss function for device $i$ is given by:
\begin{equation}
F_i(\theta) = \frac{1}{|S_i|} \sum_{(u_{i,k}, v_{i,k}) \in S_i} \mathcal{L}(\theta; u_{i,k}, v_{i,k}).
\end{equation}

Thus, the Federated Learning problem is expressed as the optimization task:
\begin{equation}
\min_\theta F(\theta) = \min_\theta \frac{1}{T} \sum_{i \in \mathcal{N}} |S_i| F_i(\theta).
\end{equation}

In synchronous federated learning, the server must wait for all chosen clients to finish their local training before aggregating the model. Conversely, AFL bypasses this constraint by instantly updating the global model whenever a client submits its update. The aggregation mechanism in AFL is described by the formula:
\begin{equation}
\mathbf{v}_k^{(m)} = \lambda \mathbf{v}^{(c)} + (1 - \lambda) \mathbf{v}_{k-1}^{(m)},
\end{equation}
where $\mathbf{v}^{(c)}$ is the client model received, $\mathbf{v}_k^{(m)}$ represents the $k$-th iteration of the global model, and $\lambda$ is determined by the staleness of the received model. Generally, a lower $\lambda$ is used for updates exhibiting notable staleness.

In semi-asynchronous federated learning, the server establishes a buffer with a size of $C$. When the buffer is filled, the system aggregates all buffered gradients to form a new gradient, which is then aggregated with the global model. Subsequently, the buffer is cleared and begins receiving gradients uploaded by clients again. The aggregation formula is as follows:
\begin{equation}
\mathbf{v}_k^{(m)} = \mathbf{v}_{k-1}^{(m)} + \frac{1}{C} \sum_{i=1}^{C} \mathbf{v}_i^{(c)} \lambda_i.
\end{equation}

Additionally, the meanings of the commonly used symbols in this paper are explained in Table \ref{tab:notations}.

\begin{figure*}[t]
    \centering
    \begin{minipage}{1.0\textwidth}
        \includegraphics[width=\linewidth]{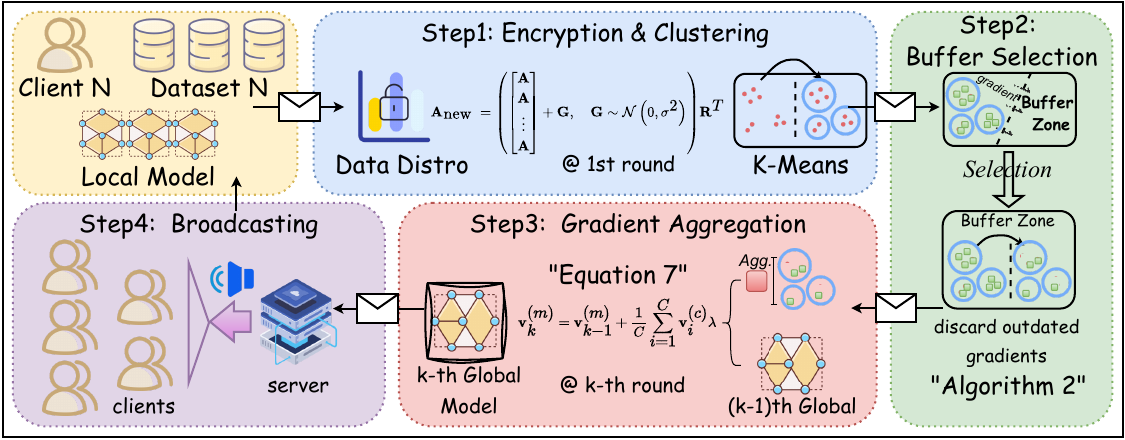}
    \end{minipage}%
    \caption{\textbf{The illustration of the proposed AFBS framework.} Clients first need to encrypt the label distribution using random projection before training. The server will cluster the clients based on this information. During training, the server collects gradients sent by clients until the buffer is full. Once the buffer is full, the server employs the AFBS algorithm to perform Gradient Selection on the gradients within each cluster. The remaining gradients in the buffer are then aggregated and combined once more with the global model. The server broadcasts this new global model to the clients, commencing a new round of federated learning.}
    \label{fig:pipeline}
\end{figure*}

% Synchronous federated learning requires waiting for all selected clients to complete their local training before the server performs model aggregation. In contrast, asynchronous federated learning overcomes this limitation by immediately aggregating the model whenever the server receives an update from any client, thereby generating an updated global model. The aggregation process in asynchronous federated learning can be expressed by the following formula:

% \begin{equation}
% \mathbf{w}_i^{(g)} = \beta \mathbf{w}^{(l)} + (1 - \beta) \mathbf{w}_{i-1}^{(g)},
% \end{equation}
% here, $\mathbf{w}^{(l)}$ represents the received client model, $\mathbf{w}_i^{(g)}$ denotes the $i$-th version of the global model, and $\beta$ depends on the value of the received model. Typically, a smaller $\beta$ is assigned to gradients with significant staleness.

\subsection{Motivation}
\begin{figure}[t]
\centering
\hspace{-0.57cm}
\includegraphics[width=1\textwidth]{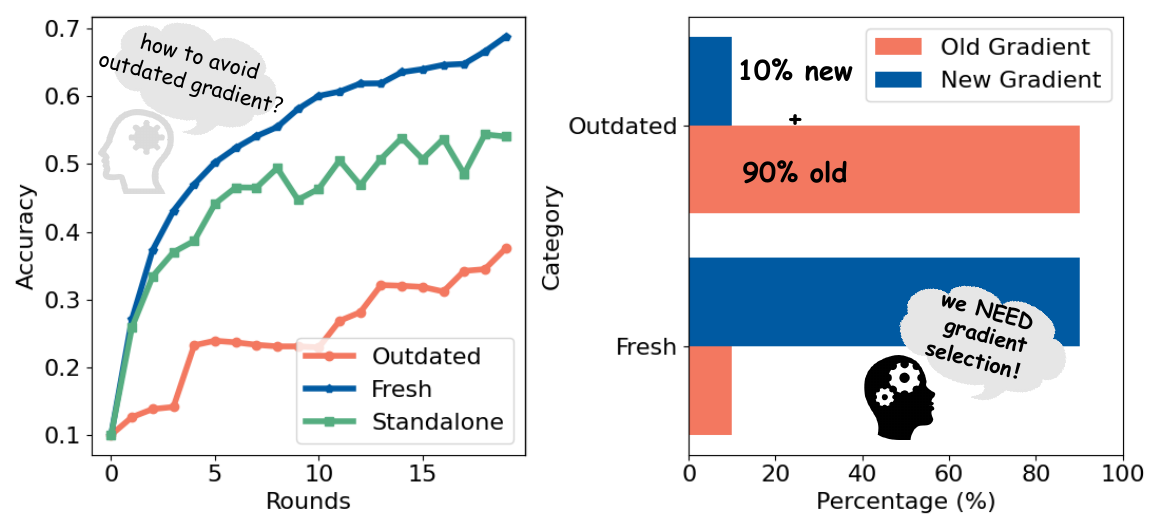}
\caption{Comparison of Aggregating Stale Gradients, Fresh Gradients, and Individual Training. The ``outdated" curve represents the aggregation of 10\% fresh gradients and 90\% outdated gradients. The ``fresh" curve represents the aggregation of 90\% fresh gradients and 10\% outdated gradients. The ``standalone" curve represents the local training of a single client.}  
\label{fig:think}
\end{figure}

The proposed method aims to mitigate the negative effects of outdated gradients in semi-asynchronous federated learning. We define the staleness of gradients as $\tau$, the difference between the current global round and the round in which the gradient source model was generated. An experiment shown in Figure \ref{fig:think} reveals that when a large number of outdated gradients are aggregated in each round, training performance significantly deteriorates, in some cases even falling below the performance of isolate single-client training. This highlights the need to remove outdated gradients from the buffer.

Due to the characteristics of AFL, the proposed method must minimize overhead while addressing the problem. Therefore, our approach should be as simple and efficient as possible. Inspired by the monotonic queue in data structures, we hypothesize that if a gradient is inferior in both staleness and data volume compared to another gradient, it can be considered negligible and removed. However, we find that this method performed poorly because data from certain categories (only present in slower clients) struggled to participate in training, leading to a reduction in the training data domain.

\begin{figure*}[t]
    \centering
    \begin{minipage}{0.32\textwidth}
        \includegraphics[width=\linewidth]{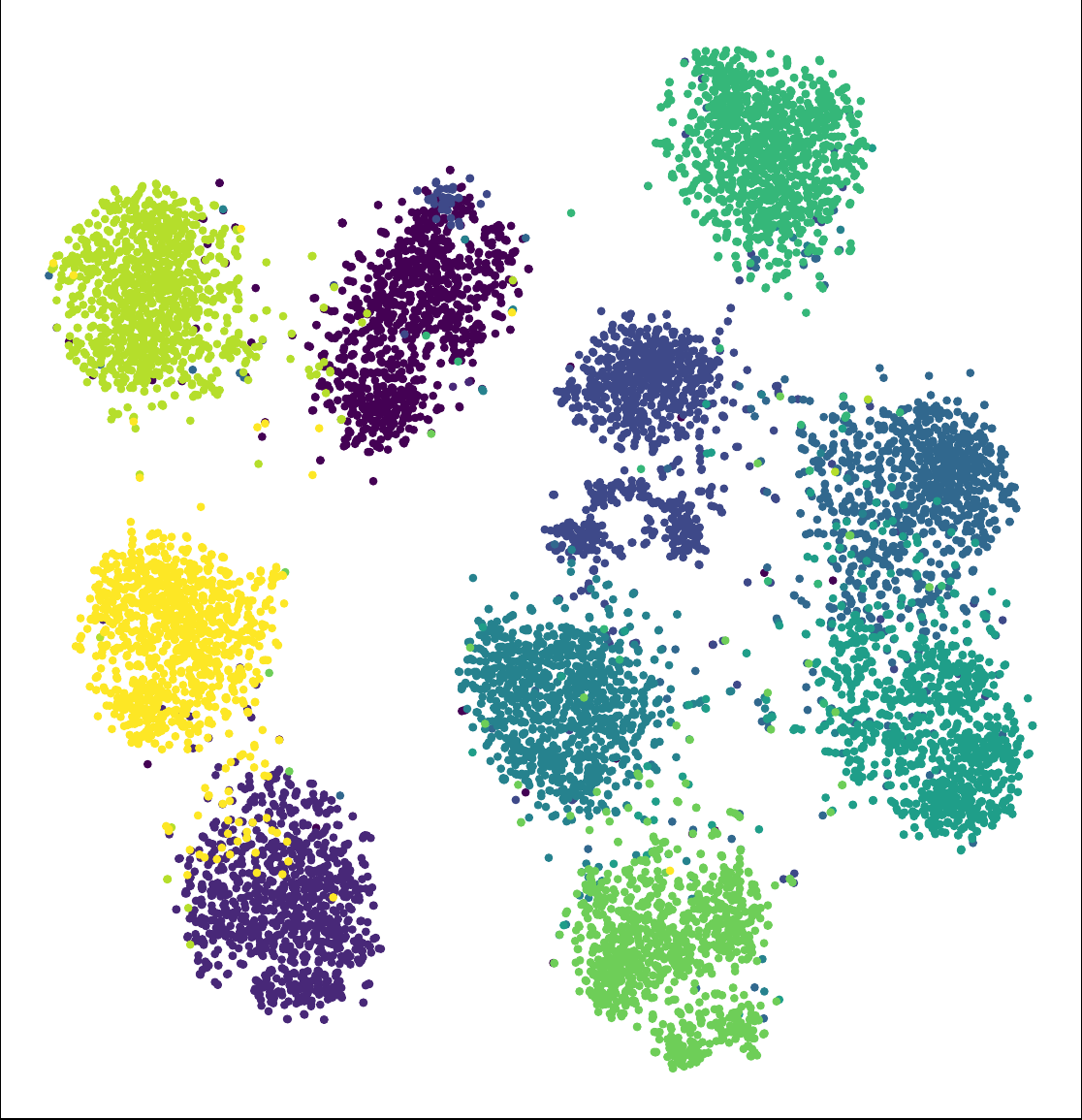}
        \subcaption{Origin} 
    \end{minipage}%
    \hspace{0.001\textwidth} 
    \begin{minipage}{0.32\textwidth}
        \includegraphics[width=\linewidth]{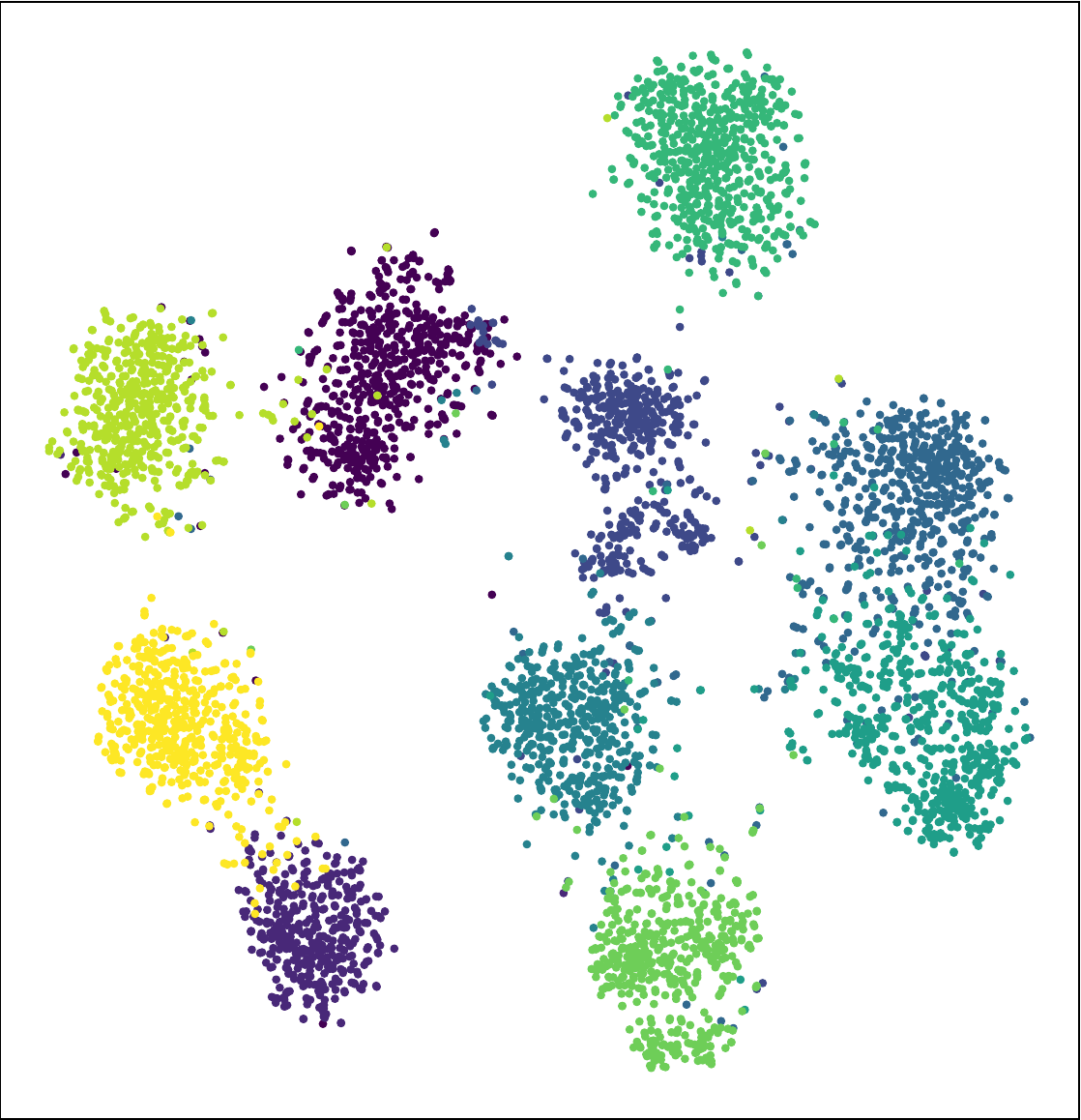}
        \subcaption{Pre-Clustered}
    \end{minipage}
    \begin{minipage}{0.32\textwidth}
        \includegraphics[width=\linewidth]{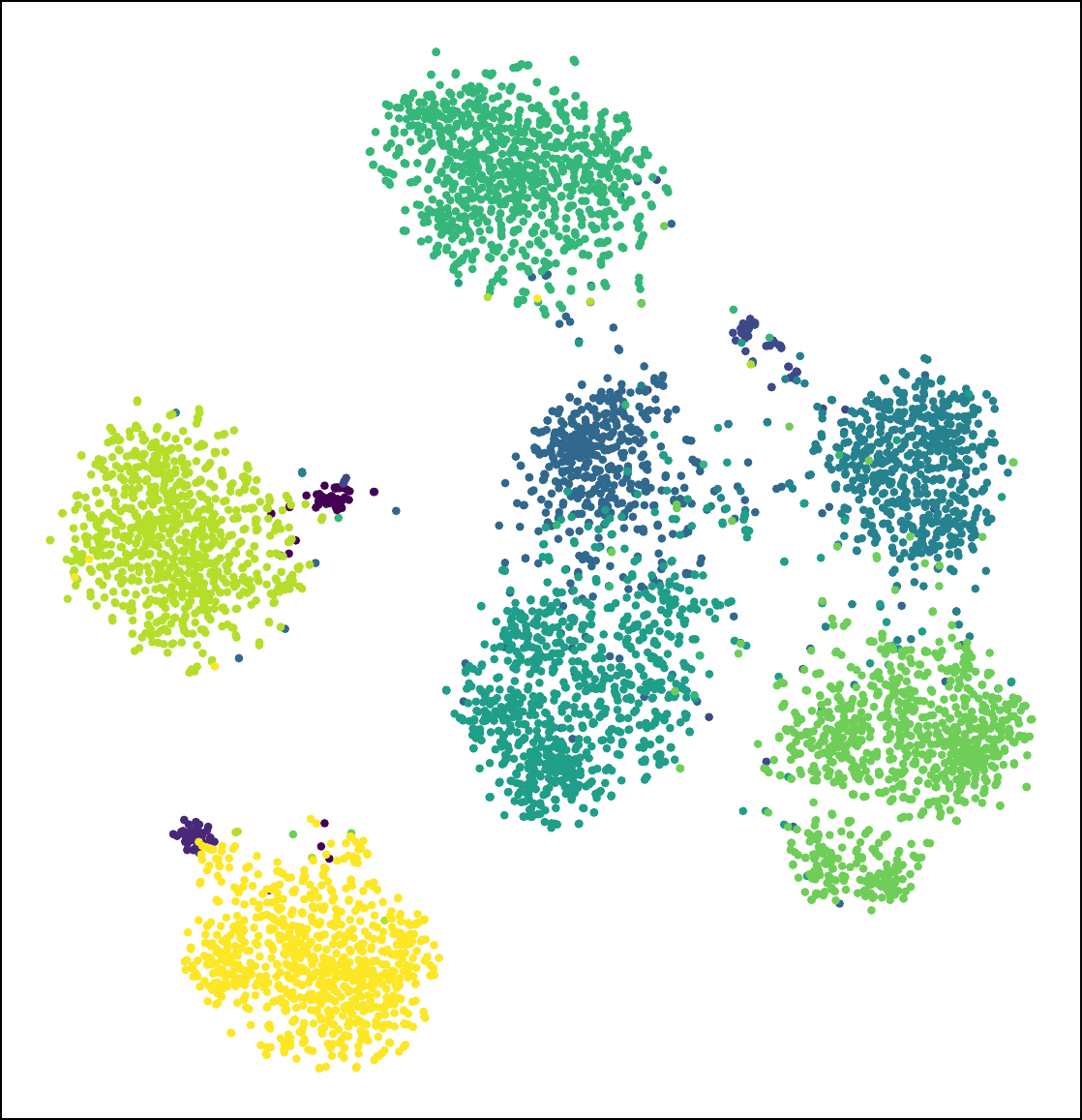}
        \subcaption{Non-Clustered}
    \end{minipage}%
    \caption{The data distribution of pre-clustering versus non-pre-clustering strategies participating in training after gradient screening. In Fig.(a), we demonstrate the complete data distribution where each color represents a class. Fig.(b) shows the data distribution participating in training after pre-clustering, where the distribution patterns of each class remain largely unchanged. Fig.(c) presents the training data distribution without pre-clustering, revealing only 7 dominant clusters compared to the original 10 classes. This indicates that strategy (c) may lead to insufficient training for certain classes processed by slow clients.}
    \label{fig:tsne}
\end{figure*}

To verify the impact of neglecting slow-labeled data without clustering, we assign a category label to each client, with increasing latency as the label number grew. We visualize the distribution of data that successfully participated in training, as shown in Figure~\ref{fig:tsne}. The results indicate that without clustering, data from certain categories may struggle to participate in training and could even be discarded, resulting in an incomplete data distribution for the final model and affecting the model’s generalization ability.

\subsection{Framework Overview}
The main focus of our algorithm is the gradient selection in the server-side buffer, with no modifications made to the client-side. The server-side operations of our algorithm are detailed in Algorithm 1, where FL-Client represents standard federated learning client training. During server-side aggregation, we perform gradient selection, and the corresponding algorithm is elaborated in Algorithm 2. The overall flowchart of the AFBS algorithm is shown in Figure \ref{fig:pipeline}.

\subsection{Client Clustering}
To avoid neglecting information from clients with different data distributions during the training process, we evaluate and select gradients only from clients with similar data distributions, which necessitates grouping the clients. To address more complex scenarios, existing methods typically group models during training by capturing the similarity between clients. These methods assume that clients with similar data distributions will exhibit higher model similarity in federated learning environments. However, this assumption may fail in AFL settings \cite{10.1145/3637528.3671979}.

Using cosine similarity as an example, we simulate the issue of staleness in AFL. In this scenario, fast-updating clients perform multiple rounds of updates, while slow clients do not update. Experimental results show that, in an asynchronous environment, the model similarity for clients with consistent data distributions may be lower. This phenomenon is in sharp contrast to synchronous federated learning, where the model similarity of clients with consistent label distributions consistently remains higher than that of clients with differing label distributions.
\begin{figure}[t]
\hspace{-1cm}
    \centering
    \begin{minipage}{0.5\textwidth}
        \includegraphics[width=\linewidth, trim=0 0 0 0, clip]{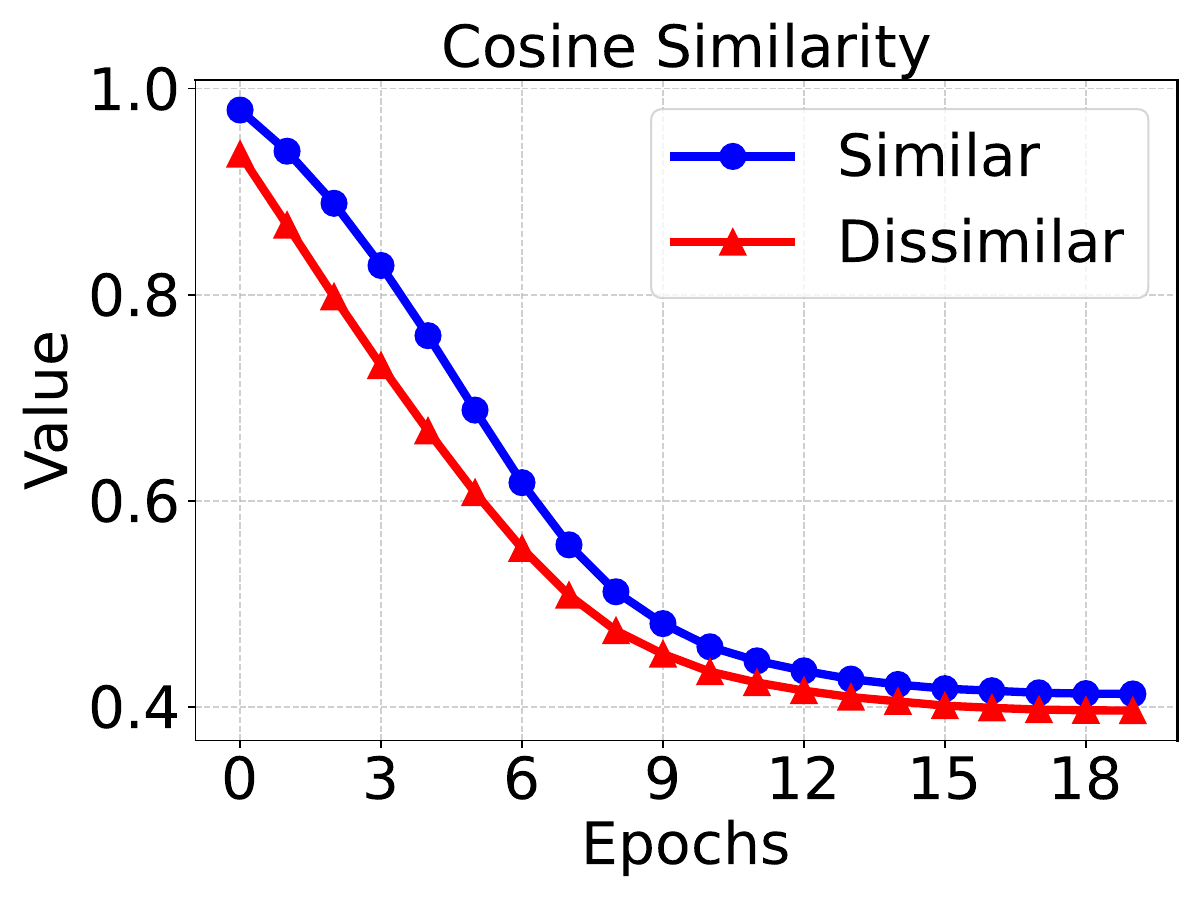}
        \subcaption{Synchronize environment}  % 子图的标题
    \end{minipage}%
    \begin{minipage}{0.5\textwidth}
        \includegraphics[width=\linewidth, trim=0 0 0 0, clip]{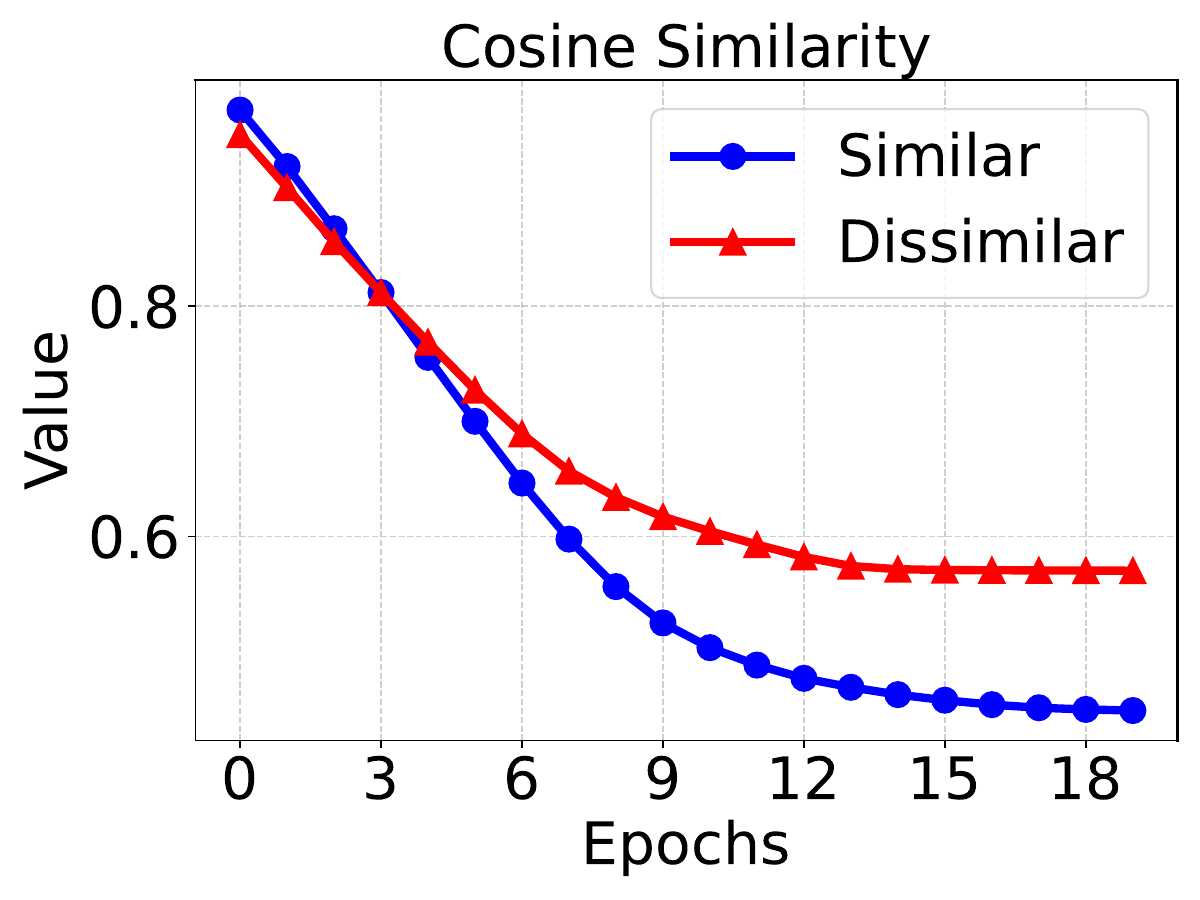}
        \subcaption{Asynchronous environment}
    \end{minipage}%
    \caption{Changes in cosine similarity of models with consistent and inconsistent label distributions under different environments.}
    \label{fig:figure4}
\end{figure}

Given the difficulty of assessing client similarity in asynchronous environments, we propose a method that incorporates a client clustering preprocessing step before the training process to mitigate this issue. Specifically, we first use random projection to encrypt the label distribution matrices of clients and then employ the K-Means algorithm \cite{5453745} to group all participating clients based on the encrypted label distribution information. This clustering strategy effectively reduces the neglect of certain data distributions during training.

\begin{algorithm}[tb]
\caption{AFBS-Server}
\textbf{Input}:  all clients $c_g$,client learning rate $\eta_c$, client SGD steps $Q$, buffer $S$, buffer size $C$, number of clusters $K$ \\
\textbf{Output}: A trained global model
\begin{algorithmic}[1]
\STATE Perform \textbf{Client Clustering} on $c_g$ to divide all clients into $K$ clusters;
\REPEAT
    \STATE $c_n \gets \text{sample available clients}$ 
    \STATE \text{Run FL-Client}$(w^t, \eta_c, Q)$ on $c_n$ 
    \IF{\text{receive client update}}
        \STATE $\Delta_i \gets \text{received update from client } i$
        \STATE Insert $\Delta_i$ into buffer $S$ 
    \ENDIF
    \IF{$\text{len}(S) \geq C$}
        \STATE $S \{\Delta_1,\dots, \Delta_n\}\gets \texttt{Gradient-Select}(S)$
        \STATE $ \bar{\Delta} = \frac{1}{|n|} \sum_{\Delta \in S} \Delta \cdot \lambda$
        \STATE $w^{t+1} \gets w^t - \eta_g \bar{\Delta}$
        \STATE Initialize buffer $S \gets \emptyset$
    \ENDIF
\UNTIL{Convergence}
\end{algorithmic}
\end{algorithm}

\subsection{Random projection encryption}
Since federated learning is privacy-preserving, we first need to encrypt the label distribution of the clients. Random projection is an excellent algorithm that maps high-dimensional data to a lower-dimensional space using a randomly generated matrix \cite{10.1145/502512.502546}.
According to the Johnson-Lindenstrauss lemma \cite{larsen2014johnsonlindenstrausslemmaoptimallinear}, we can obtain the following inequality, where $p$ and $q$ are two points in space:
\begin{equation}
    (1 - \epsilon)\|p - q\|^2 \leq \|\mathbf{R}p - \mathbf{R}q\|^2 \leq (1 + \epsilon)\|p - q\|^2.
\end{equation}

This ensures that the distance between point pairs in high-dimensional space in the algorithm can still remain roughly unchanged in low-dimensional space.
Let us denote the label distribution matrix as $\mathbf{A} \in \mathbb{R}^{1 \times d}$, and we perform the following operations to obtain a new matrix, where $\sigma$ is a specified small constant.

\begin{equation}
\mathbf{A}_{\text{new}} = 
\left(
\begin{bmatrix}
\mathbf{A} \\
\mathbf{A} \\
\vdots \\
\mathbf{A}
\end{bmatrix}
+ \mathbf{G}, \quad \mathbf{G} \sim \mathcal{N}(0, \sigma ^ 2
)
\right) \mathbf{R}^T.
\end{equation}

We replicate the original matrix to generate a new matrix of size \( d \times d \). On this basis, an additional noise matrix following a Gaussian distribution is added to ensure that the rank of the new matrix reaches \( d \). Due to the independence of the Gaussian random distribution and the properties of matrix addition, we have the following conclusion:
\begin{equation}
\mathbb{P}(\det(A+G)=0)=0,
\end{equation}
where $A$ is an arbitrarily given matrix, and $G$ is a Gaussian noise matrix. The detailed proof will be provided in the Supplementary. This approach guarantees that subsequent random projection operations will necessarily reduce the rank of the matrix, effectively preventing the original matrix \( \mathbf{A} \) from being decrypted from the generated new matrix \( \mathbf{A}_{\text{new}} \). Each client sends \( \mathbf{A}_{\text{new}} \) to the server. Due to the ability of random projection to approximately preserve pairwise distances, we can still effectively perform clustering on the label distribution.

\subsection{AFBS}
Our method scores gradients based on their staleness and the data size of the client to which the gradient belongs, aiming to eliminate uninformative gradient information in the buffer.

\begin{algorithm}[tb]
\caption{Gradient-Select}
\label{alg:select}
\textbf{Input}: buffer $S$, number of clusters $K$\\
\textbf{Output}: Filtered buffer
\begin{algorithmic}[1]
\STATE $S_k = \{ x \in S \mid \text{Cluster}(x) = k \}, \quad k = 1, 2, \dots, K$
\FOR{each $S_k$}
    \FOR{each sample $x \in S_k$}
        \STATE $score(x) \gets \text{Score}(x)$
    \ENDFOR
    \STATE $x_{\text{max}} \gets \arg\max_{x \in S_k} score(x)$ 
    \STATE $\mathcal{V}_{\text{m}} \gets \mathcal{V}(x_{\text{max}})$
    \STATE $\tau_m \gets \tau(x_{\text{max}})$
    \STATE $S_k \gets \{ x \in S_k \mid \mathcal{V}_x \geq \mathcal{V}_m \lor \tau_x \leq \tau_m \}$
\ENDFOR
\STATE $S = \bigcup_{k=1}^K S_k$
\end{algorithmic}
\end{algorithm}

The scoring function is defined as follows:
\begin{equation}
\text{Score}(x) = \frac{\mathcal{V}_{\text{x}}}{(\tau_x + 1)^2}.
\end{equation}

Once the gradients in the buffer reach the specified value $C$, we first group the gradients based on the clients to which they belong, and then perform gradient selection within each group based on a scoring function. Specifically, we attempt to eliminate all gradients whose staleness and data size are worse than the highest-scoring gradient. However, to avoid getting trapped in a local optimum, we do not eliminate all the seemingly inferior gradients. Instead, we perform a probabilistic selection based on the scores, with the probability of a gradient being selected given by:

\begin{equation}
p(i) =  \frac{\text{score}(x_i)}{\|\text{score}(x_j)\|_{\infty}}, \quad x_j \in S.
\end{equation}

\section{Experiment}
In this paper, we conduct experiments using four different datasets and compare our method with six other federated learning algorithms. The results demonstrate that, under various experimental settings, the performance of our method is superior to the others, and it has a significant advantage in more challenging tasks.

\subsection{Experimental Setup}
We implement AFBS using PyTorch and FLGO \cite{wang2023flgofullycustomizablefederated}, FLGO is a federated learning framework that supports AFL through time simulation. Our experiments are conducted on NVIDIA GeForce RTX 4090 GPU. 

\textbf{Settings. }   For computer vision (CV) tasks, we simulate a federated learning scenario with 600 clients, sampling 120 clients per training round and running for 20 virtual days. For natural language processing (NLP) tasks, we use 60 clients, sampling 12 per round and running for 10 virtual days. Each client's latency is uniformly distributed between 0 and 6000 seconds.

\textbf{Hyperparameters. } All experiments use a learning rate of 0.01 with a decay factor of 0.999, 5 local epochs, and a batch size of 64. For the asynchronous algorithm with a buffer, the buffer size is set to 10. In the AFBS algorithm, the obsolescence decay factor $\lambda$ is $\frac{1}{\sqrt{\tau^{\text{min}} + 1}}$, and the $\sigma$ is set to 1 $\times 10^{-3}$.

\textbf{Datasets and Models. } We test four datasets in total, covering both CV and NLP tasks. For CV, we test the LeNet5  model \cite{726791} on MNIST \cite{lecun2010mnist}, CIFAR-10~\cite{Krizhevsky09learningmultiple}, and CIFAR-100 \cite{Krizhevsky09learningmultiple}. For NLP, we test the FastText model \cite{joulin2016bagtricksefficienttext} on SST2~\cite{xiao2017/online}. We partition the dataset into K clusters using a Dirichlet distribution with $\alpha$ equal to 0.1. Within each cluster, the data distribution is independent and identically distributed, and the data volume follows a log-normal distribution with a standard deviation of 1.

\begin{table*}[t]\tiny
\centering
\caption{The highest accuracy achieved by different algorithms on different datasets within a fixed virtual runtime. The numbers in bold represent the highest values, and the underlined numbers represent the second-highest values.}
\label{tab:highresults}
\resizebox{\textwidth}{!}{%
\begin{tabular}{lccc|ccc|ccc|cc}
\toprule
\multirow{3}{*}{\textbf{Methods}} & \multicolumn{3}{c|}{\textbf{MNIST $\uparrow$}} & \multicolumn{3}{c|}{\textbf{CIFAR-10 $\uparrow$}} & \multicolumn{3}{c|}{\textbf{CIFAR-100} $\uparrow$} & \multicolumn{2}{c}{\textbf{SST2 $\uparrow$}} \\ 
\cmidrule(r){2-12}
                 & K=1            & K=3         & K=5            & K=1          & K=3            & K=5          & K=1            & K=3          & K=5            & K=1          & K=2        \\ \midrule
FedAVG           & 98.56           & 97.39           & 96.05           & 59.11           & 52.74           & 44.48           & 20.31           & 17.82           & 15.02           & 62.50           & 64.79                 \\
FedAsync           & 85.16           & 83.82           & 80.76           & 17.01           & 19.33           & 19.88           & 1.15           & 1.13           & 1.04           & 58.83           & 59.63                   \\
FedBuff          & 98.26           & 97.79           & 97.13           & 56.73           & 50.05           & 47.07           & 18.76           & 17.24           & 16.87           & 62.39           & 62.96                    \\
CA2FL           & \underline{98.83}           & \underline{98.74}           & \textbf{98.72}           & \underline{61.53}           & \underline{57.52}           & \underline{54.85}           & 
25.18           & 19.42           & 17.66           & 62.27           & 64.11                   \\
FedFa           & 79.92           & 81.99           & 76.60           & 14.49           & 13.32           & 19.15           & 1.15           & 1.15           & 1.15           & 57.45           & 57.91                     \\
FedDyn           & 98.09           & 98.08           & 98.42           & 54.59           & 50.80           & 51.37           & \underline{25.20}           & \underline{21.10}           & \underline{20.30}            & \textbf{65.48}           & \textbf{67.89}                   \\
% AFBS-JD           & 98.69           & 98.72           & 98.56           & 62.79           & 58.13           & 55.30           &25.08           &22.02           &21.15            &-           &-                   \\
\rowcolor[HTML]{ededed}
\textbf{AFBS}           & \textbf{98.91}  & \textbf{98.77}  & \underline{98.62}  & \textbf{65.79}  & \textbf{61.85}  & \textbf{55.83}  & \textbf{30.00}  & \textbf{24.36}  & \textbf{22.93}  & \underline{63.99}  &\underline{65.25}    \\ 
\bottomrule
\end{tabular}%
}
\end{table*}

\begin{table}[t]
\centering
\caption{Final accuracy after 20 days of training.}
\label{tab:finalresults}
\resizebox{0.5\columnwidth}{!}{%
\begin{tabular}{lcc|cc|cc}
\toprule
\multirow{2}{*}{\textbf{Strategy}} & \multicolumn{2}{c|}{\textbf{MNIST $\uparrow$}} & \multicolumn{2}{c|}{\textbf{CIFAR-10 $\uparrow$}} & \multicolumn{2}{c}{\textbf{CIFAR-100 $\uparrow$}}\\ 
\cmidrule(r){2-7}
                 & K=3         & K=5           & K=3            & K=5         & K=3          & K=5       \\ \midrule
w/o clustering           & 98.56           & 98.20           & 57.21           & 50.27           & 21.32           & 20.24       \\
\rowcolor[HTML]{ededed}
w/ clustering          & \textbf{98.72}           & \textbf{98.61}           & \textbf{61.65}           & \textbf{55.51}           & \textbf{24.15}           & \textbf{22.58}                      \\
\bottomrule
\end{tabular}%
}
\end{table}

\textbf{Baselines.} To validate the performance and applicability of our proposed algorithm in asynchronous environments, we conduct a comparative study with several classic and state-of-the-art federated learning algorithms on these datasets. These included FedAVG \cite{pmlr-v54-mcmahan17a} and FedDyn \cite{acar2021federatedlearningbaseddynamic} as synchronous algorithms, FedAsync~\cite{xie2020asynchronousfederatedoptimization} and FedFa \cite{10.24963/ijcai.2024/584} as asynchronous algorithms, and FedBuff \cite{nguyen2022federatedlearningbufferedasynchronous} and CA2FL~\cite{wang2024tackling} as semi-asynchronous algorithms. FedAVG and FedDyn focus on achieving stable model updates through periodic global aggregation. FedAsync and FedFa reduce communication latency with asynchronous client updates but may cause instability in model convergence. FedBuff and CA2FL combine synchronous and asynchronous strategies using buffer mechanisms or dual-layer clustering, enhancing update efficiency and improving model stability and robustness. These experiments demonstrate the superior performance and applicability of our proposed AFBS algorithm in AFL scenarios.
\begin{table*}[t]
\centering
\caption{Virtual time required for different methods to achieve target accuracy. `-' represents failure to achieve accuracy. The numbers in bold represent the lowest values.}
\label{tab:timeresults}
\resizebox{\textwidth}{!}{%
\begin{tabular}{lccc|ccc|ccc|cc}
\toprule
\multirow{3}{*}{\textbf{Methods}} & \multicolumn{3}{c|}{\textbf{MNIST $\downarrow$}} & \multicolumn{3}{c|}{\textbf{CIFAR-10 $\downarrow$}} & \multicolumn{3}{c|}{\textbf{CIFAR-100 $\downarrow$}} & \multicolumn{2}{c}{\textbf{SST2 $\downarrow$}} \\ 
                 & \multicolumn{3}{c|}{\textit{Target Acc: 0.90}} & \multicolumn{3}{c|}{\textit{Target Acc: 0.50}} & \multicolumn{3}{c|}{\textit{Target Acc: 0.15}} & \multicolumn{2}{c}{\textit{Target Acc: 0.57}} \\
\cmidrule(r){2-12}
                 & K=1            & K=3         & K=5            & K=1          & K=3            & K=5          & K=1            & K=3          & K=5            & K=1          & K=2        \\ 
\midrule
FedAVG           & 59470           & 196603           & 375483           & 810776           & 1179958           &-           & 971583        & 1191830          & 1363839           & 164930           & 179733                \\
FedAsync         & -               & -           & -           & -           & -           & -           & -            & -           & -           & 206300           &178906                   \\
FedBuff          &23137           & 46652           & 78292           & 234115           & 768587           & -           & 399040           & 509801           & 619446           & 156365           & 137166                    \\
CA2FL            & 17368           & 32695           & 95277           & 300058           & 488623           & 698699           & 300248               & 511149          & 743356          & 132056           & 163681                  \\
FedFa            & -           & -           & -           & -           & -           & -           & -            & -           & -           & 457596           & 219700                     \\
FedDyn           & 95236           & 154776           & 196603           & 518697           & 1334134          & 1322022           & 554513          & 768899        & 888089           & 179733  &174432          \\
\rowcolor[HTML]{ededed}
\textbf{AFBS}             & \textbf{16130}  & \textbf{31264}  & \textbf{52897}  & \textbf{101582}  & \textbf{237562}  & \textbf{590471}  & \textbf{138274}  & \textbf{275066}  & \textbf{393576}  & \textbf{108551}   & \textbf{130854}    \\ 
\bottomrule
\end{tabular}%
}
\end{table*}

\begin{figure*}[t]
\hspace{-0.2cm}
    \centering
    \begin{minipage}{0.45\textwidth}
        \includegraphics[width=\linewidth, trim=0 0 0 0, clip]{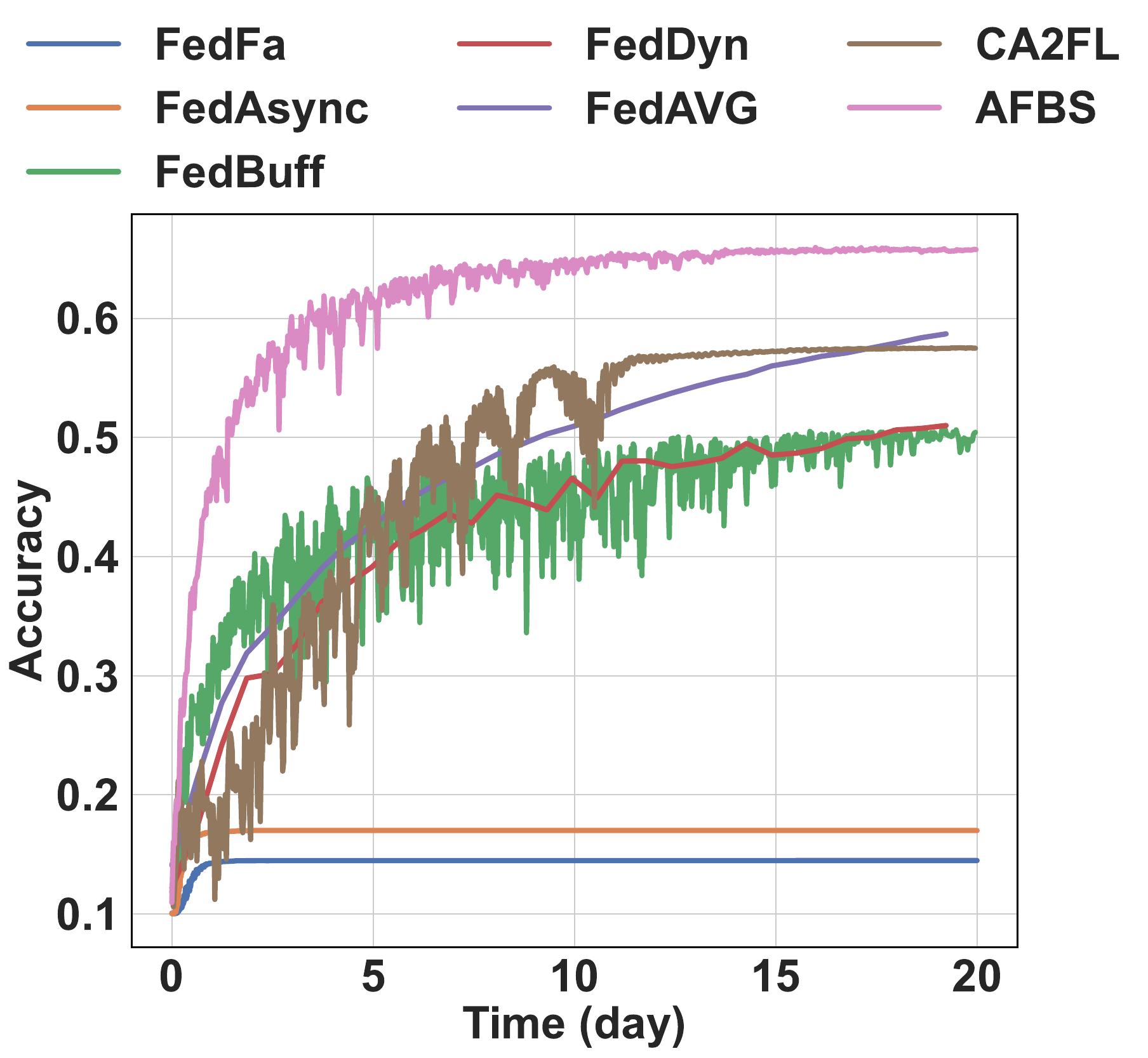}
        \subcaption{CIFAR-10}  % 子图的标题
    \end{minipage}%
    \begin{minipage}{0.45\textwidth}
        \includegraphics[width=\linewidth, trim=0 0 0 0, clip]{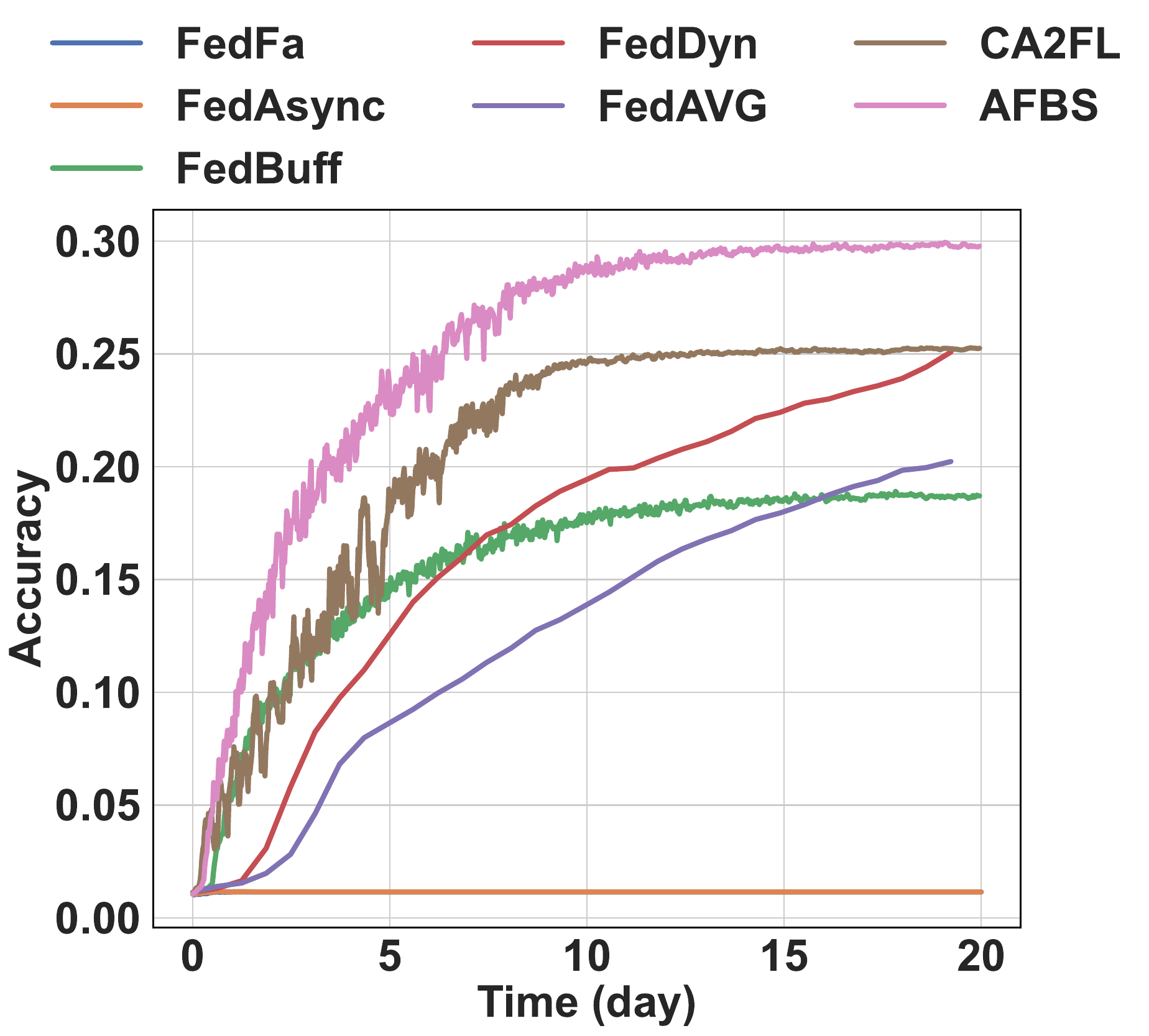}
        \subcaption{CIFAR-100}
    \end{minipage}%
    \\
    \begin{minipage}{0.45\textwidth}
        \includegraphics[width=\linewidth, trim=0 0 0 0, clip]{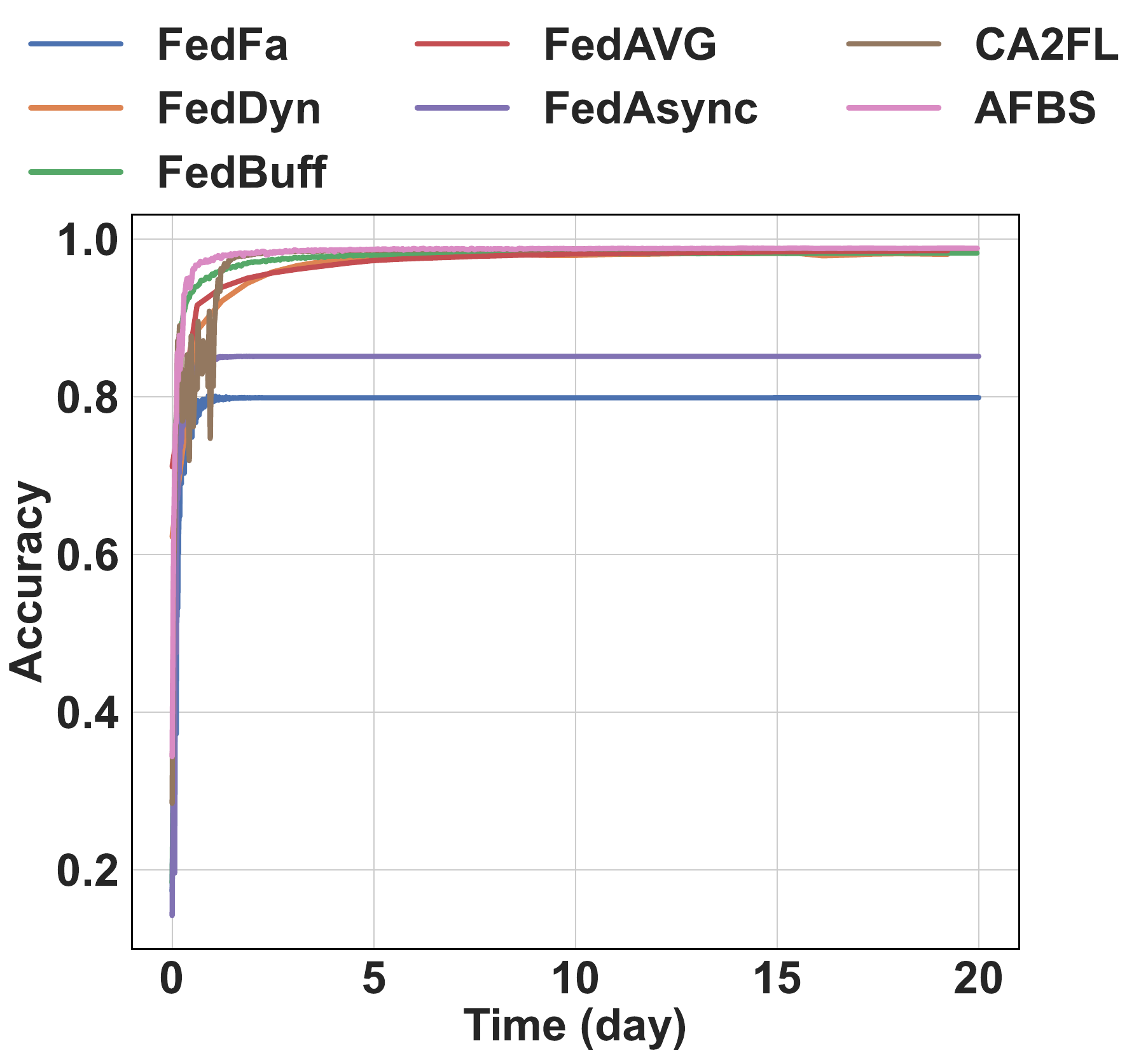}
        \subcaption{MNIST}  % 子图的标题
    \end{minipage}%
    \begin{minipage}{0.45\textwidth}
        \includegraphics[width=\linewidth, trim=0 0 0 0, clip]{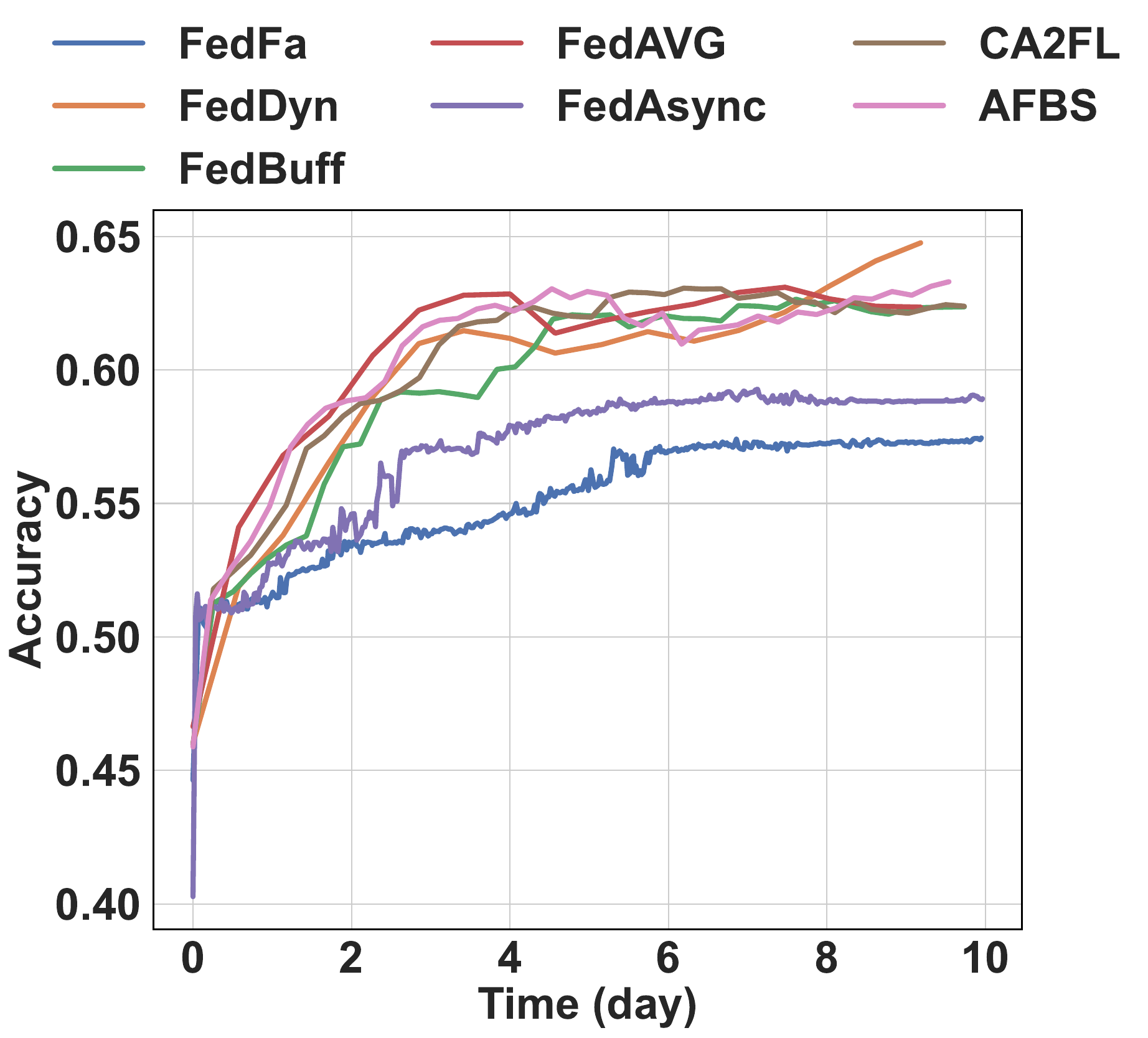}
        \subcaption{SST2}
    \end{minipage}%
    \caption{
Comparison of different algorithms on different datasets when K = 1.}
    \label{fig:figure5}
    \vspace{-0.2cm}
\end{figure*}

\subsection{Performance Comparison}
We conduct experiments on all baseline methods under the condition of fixed virtual runtime and record the highest accuracy achieved by each algorithm, as shown in Table~\ref{tab:highresults}. It indicates AFBS outperforms other methods significantly in terms of accuracy on the two most challenging datasets, CIFAR-10 and CIFAR-100. As shown in Figure~\ref{fig:figure5}, our algorithm consistently outperforms the runner-up algorithm throughout the training process, with a leading margin of \textbf{4.26\%} and \textbf{4.8\%} respectively. Additionally, in the NLP task, the accuracy of AFBS is only slightly lower than FedDyn. This is because the number of clients deployed in the NLP task is only one-tenth of that in the CV task, making the impact of staleness relatively smaller in the NLP task. Nevertheless, even in this scenario, our method still outperforms all asynchronous and semi-asynchronous algorithms in terms of accuracy. The above experimental results fully demonstrate that the proposed AFBS method exhibits outstanding performance under different levels of staleness impact.

It is noteworthy that FedFa's performance is unexpectedly low. Through experimental validation, we attribute this to the long-tail distribution of response times adopted in the FedFa paper. Under that experimental setup, the vast majority of devices exhibit extremely short response times, while only a very small fraction have significantly longer delays. In contrast, our experiments employ a uniform distribution of response times,where all devices' delays are evenly distributed within a specified interval. This setup is more challenging and better aligns with real-world scenarios.

\subsection{Convergence Effectiveness}
We analyze the virtual time required for different baseline methods to achieve the target accuracy, with the specific results shown in the Table \ref{tab:timeresults}. Since the primary goal of asynchronous algorithms is to be fast and efficient, \textit{i.e.}, to achieve the target accuracy in the shortest possible time, this metric is of great importance. From the results, it can be observed that the time required by our method to reach the target accuracy in all tasks is significantly lower than that of all synchronous federated learning algorithms and also superior to all AFL algorithms, which fully demonstrates the efficiency of our method.

\begin{figure}[t]
\hspace{-0.3cm}
    \centering
    \begin{minipage}{0.5\textwidth}
        \hspace{-0.2cm}
        \includegraphics[width=\linewidth]{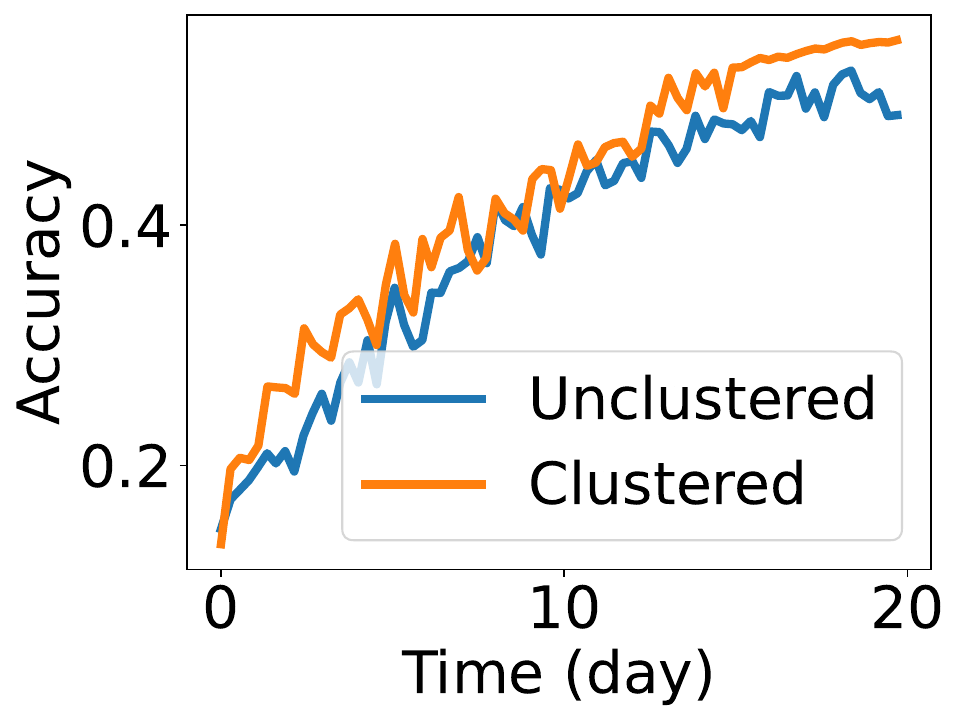}
        \subcaption{CIFAR-10}  % 子图的标题
    \end{minipage}%
    \begin{minipage}{0.5\textwidth}
        \hspace{-0.2cm}
        \includegraphics[width=\linewidth]{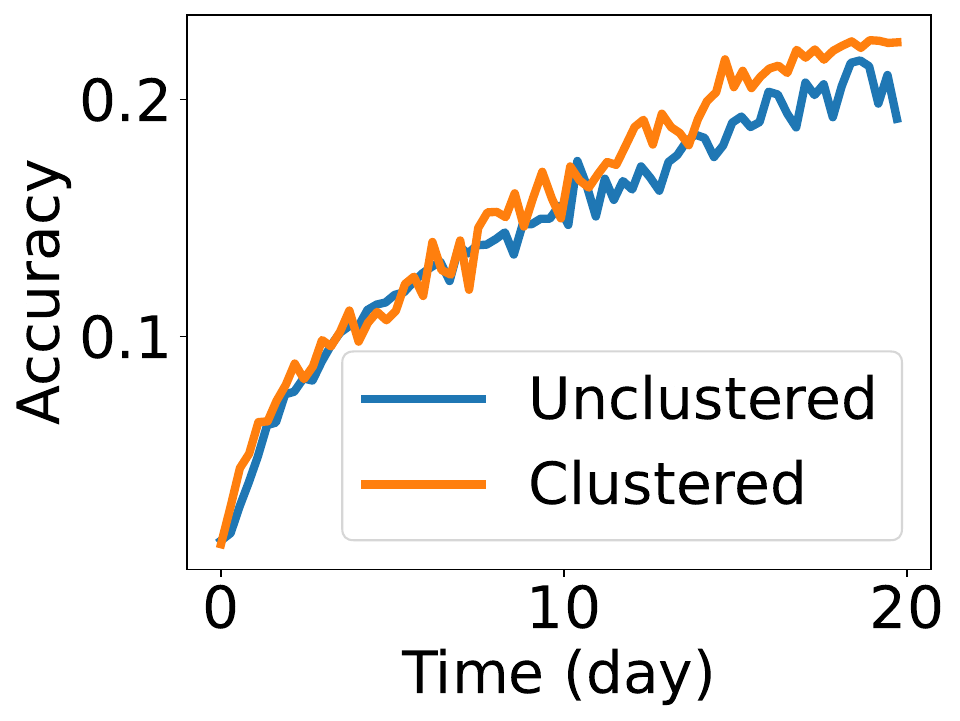}
        \subcaption{CIFAR-100}
    \end{minipage}%
    \caption{
Comparison results of clustering or not.}
    \label{fig:figure6}
    \vspace{-0.3cm}
\end{figure}

\subsection{Effectiveness of Clustering}
Table \ref{tab:finalresults} shows the impact of clustering on algorithm performance. When the data distribution across clients is heterogeneous, skipping the clustering process and directly applying the gradient selection algorithm to the entire buffer leads to a decline in the final model accuracy. This decline primarily stems from the poor performance of certain client categories, where the algorithm tends to discard the gradients generated by these categories. As a result, the model struggles to effectively learn the features of these categories, thereby reducing overall accuracy.
% However, we also observed that during training, the model performance without clustering occasionally surpasses that with clustering. This phenomenon is attributed to the excessive elimination of gradients, which allows clients with stronger performance to dominate the training process, temporarily improving the model's performance.

\subsection{Effectiveness of Buffer Size}
\begin{figure}[t]
\hspace{-0.3cm}
    \centering
    \begin{minipage}{0.5\textwidth}
        \hspace{-0.2cm}
        \includegraphics[width=\linewidth]{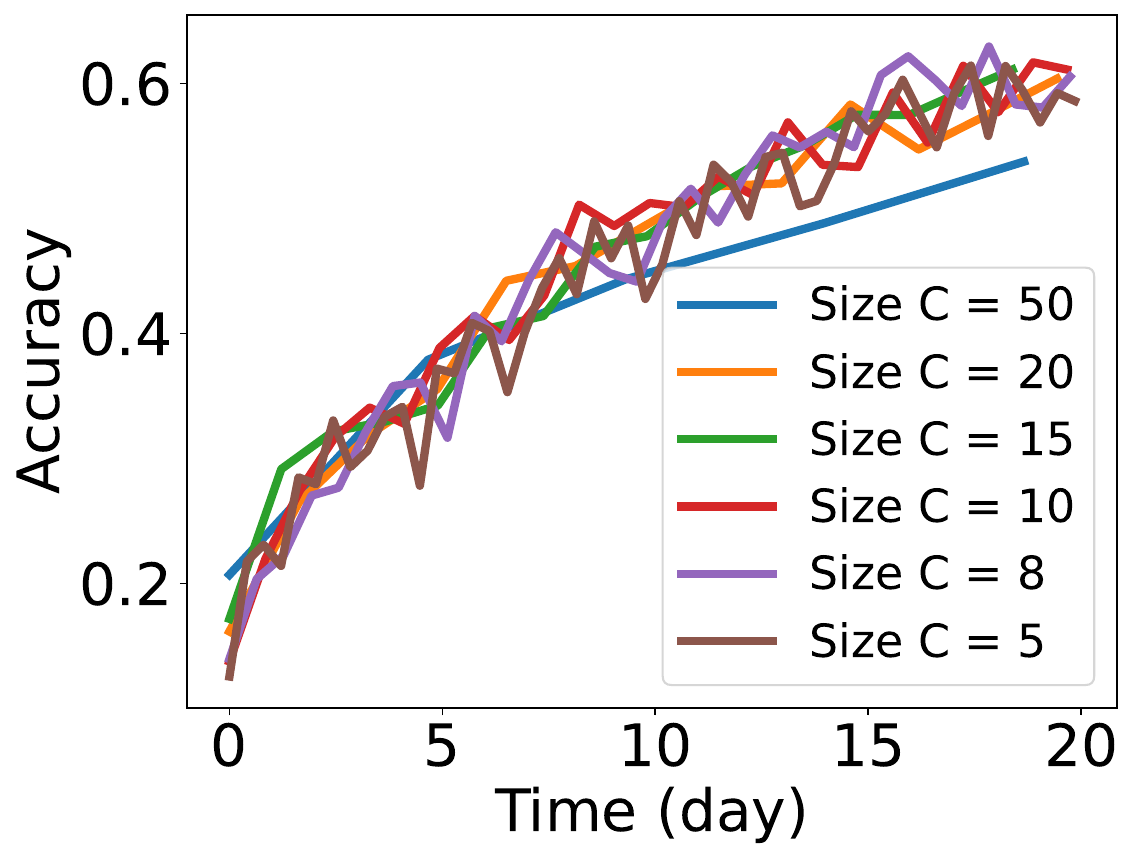}
        \subcaption{K = 5}  
    \end{minipage}%
    \begin{minipage}{0.5\textwidth}
        \hspace{-0.2cm}
        \includegraphics[width=\linewidth]{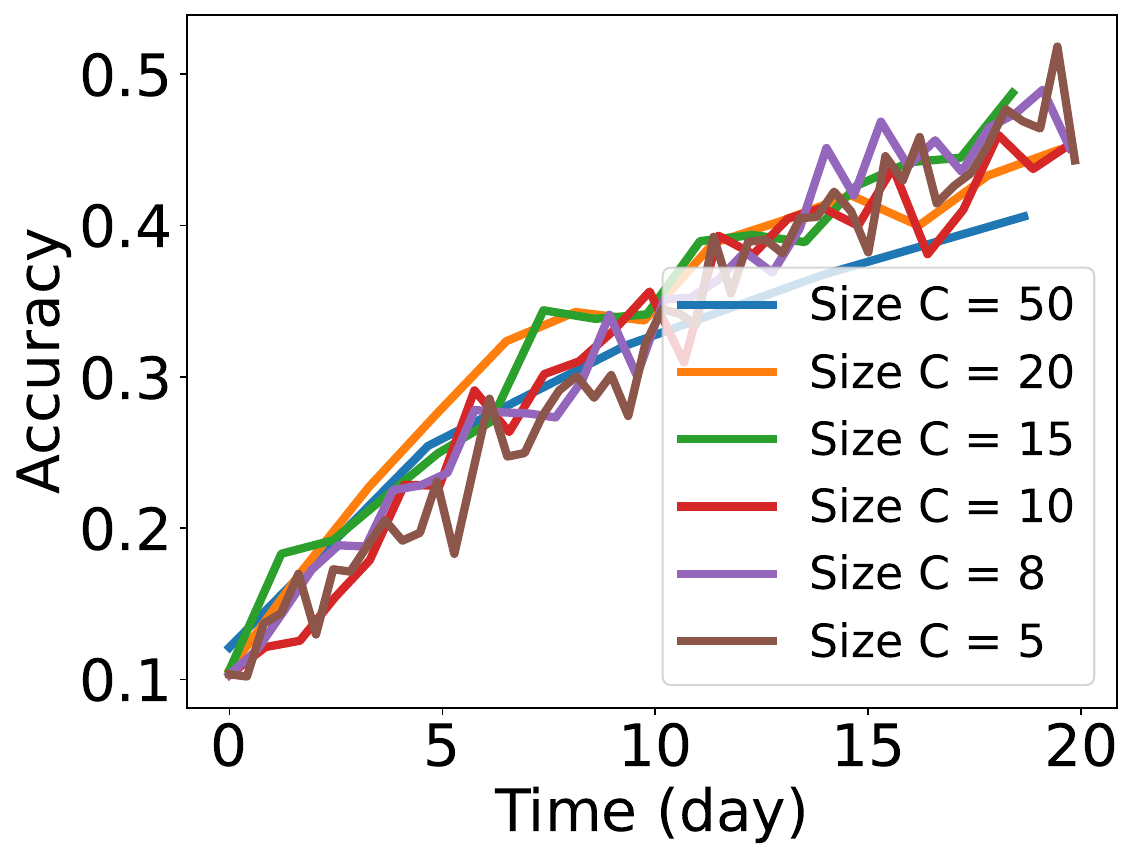}
        \subcaption{K = 10}
    \end{minipage}%
    \caption{
Impact of buffer size under different K.}
    \label{fig:figure7}
    \vspace{-0.3cm}
\end{figure}

We evaluate the effect of buffer size $C$ on the algorithm's final accuracy using the CIFAR-10 dataset, as shown in Figure~\ref{fig:figure7}. It shows that adjusting $C$ within a reasonable range has minimal impact on model performance, indicating that our algorithm is robust to buffer size. Increasing $C$ slows aggregation but helps the algorithm discard outdated gradients, stabilizing performance. However, when $C$ = 50, convergence speed significantly decreases, as a larger $C$ makes the semi-asynchronous algorithm behave more synchronously, losing its speed advantage. Additionally, we do not recommend setting $C$ to a very small value, as this may result in too few gradients with the same distribution in the buffer, failing to fully leverage the gradient selection method's effectiveness.

\subsection{Scalability in the Face of Limited Resources}
The core operation of the proposed method is gradient selection performed at the server side. While this strategy effectively enhances model performance, the additional computational overhead may imposes significant pressure on server resources. Particularly in asynchronous environments, the server's performance faces even greater challenges, which limits the algorithm's applicability in resource-constrained scenarios. To systematically evaluate the scalability of the AFBS algorithm in low-resource environments, we conduct experimental analyses of the computational overhead across different application scenarios. The experimental results are shown in the Table \ref{tab:resource}. We compare the time consumption of different algorithms within a single round to quantify the additional computational costs introduced by various algorithms. Surprisingly, our algorithm introduces the lowest additional time overhead, even outperforming the baseline FedBuff algorithm.

Contrary to the hypothesis that gradient selection introduces computational overhead, our experimental results reveal that the proposed algorithm achieves a 10\%-30\% reduction in computational cost over the FedBuff baseline. Through analysis, we find that the gradient selection operation decreases the number of gradients in the buffer, which in turn reduces the quantity of gradients involved in aggregation. Since gradient summation is highly time-consuming, our method effectively lowers computational expenses. In contrast, the extra operations introduced by our algorithm only involve simple arithmetic computations within the buffer's capacity, rendering them negligible compared to gradient operations. Therefore, our algorithm demonstrates excellent performance under low-resource conditions.

\begin{table}[ht]
\centering
\caption{Runtime comparison of algorithms per round on multiple datasets. Handle Time refers to the time spent by the algorithm on the server side, while Train Time represents the duration from when the buffer is empty until it becomes full.}
\begin{tabular}{l c c c c}
\toprule
\multirow{2}{*}{Dataset} & \multirow{2}{*}{Train Time (ms)} & \multicolumn{3}{c}{Handle Time (ms)} \\
\cmidrule(lr){3-5}
& & CA2FL & FedBuff & \textbf{AFBS} \\
\midrule
MNIST    & 3764 & 140.75 & 41.16 & \textbf{36.02} \\
CIFAR-10  & 4246 & 102.06 & 24.56 & \textbf{19.58} \\
CIFAR-100 & 4724 & 79.40 & 26.94 & \textbf{19.56} \\
SST2     & 2974 & 810.71 & 275.44 & \textbf{211.45} \\
\bottomrule
\end{tabular}
\label{tab:resource}
\end{table}

\section{Conclusion}
In this paper, we propose AFBS, a gradient selection algorithm with privacy protection for semi-asynchronous federated learning. AFBS leverages a label distribution matrix after random projection to divide all clients into several clusters and eliminates the gradients with poorer performance within the same cluster in the buffer. Compared to methods addressing the staleness problem in AFL, AFBS fully utilizes the advantages of the buffer by grouping and scoring the gradients, ultimately discarding low-value gradients rather than simply reducing their weights, it leads to faster training speeds and lower computational costs. Under our experimental settings, AFBS demonstrates significant advantages in terms of final accuracy, the time required to converge to the target accuracy, and the highest achieved accuracy, especially excelling in highly challenging tasks. Our experiments prove that removing low-value gradients within the buffer can effectively enhance the performance of semi-asynchronous federated learning. However, as the first algorithm to perform gradient selection within the buffer, our method still has its shortcomings, such as performing exceptionally well only under the data distribution of personalized federated learning. We hope that there will be more in-depth research in this field in the future, which can more accurately capture the potential relationships of gradients within the buffer under asynchronous environments, thereby leading to even better algorithms.

\section{Supplementary}
\section*{Proof of Convergence Analysis}
\subsection*{\textbf{Notation}}
\begin{itemize}[leftmargin=*]
    \item $[m]$ denotes all the indices of the clients.
    \item $f^*$ denotes the optimal value of $f(w)$.
    \item $g_i(w; \zeta_i)$ represents the stochastic gradient for client $i$.
    \item $C^t$ is the buffer length after gradient selection algorithm.
\end{itemize}

\subsection*{\textbf{Assumptions}}
We retain the following assumptions:

\begin{assumption}[Gradient Unbiasedness]
\label{assumption:unbiased}
\begin{equation}
\mathbb{E}_{\zeta_i}[g_i(w; \zeta_i)] = \nabla F_i(w).
\end{equation}
\end{assumption}

\begin{assumption}[Bounded Variance]
\label{assumption:variance}
For all clients $i \in [m]$,
\begin{equation}
\mathbb{E}_{\zeta_i}\bigl[\|g_i(w; \zeta_i) - \nabla F_i(w)\|^2\bigr] \leq \sigma_\ell^2,
\end{equation}
and for global,
\begin{equation}
\frac{1}{m} \sum_{i=1}^m \|\nabla F_i(w) - \nabla f(w)\|^2 \leq \sigma_g^2.
\end{equation}

\end{assumption}

\begin{assumption}[Bounded Gradient]
\label{assumption:bounded_gradient}
\begin{equation}
\|\nabla F_i(w)\|^2 \leq G \quad \text{for } i \in [m].
\end{equation}

\end{assumption}

\begin{assumption}[Lipschitz Gradient]
\label{assumption:lipschitz}
For all clients $i \in [m]$, the gradient is $L$-smooth, i.e.,
\begin{equation}
\|\nabla F_i(w) - \nabla F_i(w')\|^2 \leq L^2 \|w - w'\|^2.
\end{equation}
\end{assumption}

\begin{assumption}[Bounded Buffer Length]
\label{assumption:buffer_size}
The buffer length \( C^t \) satisfies
\begin{equation}
\mathcal{C}_{\text{min}} \leq C^t \leq \mathcal{C}_{\text{max}},
\end{equation}
\end{assumption}
\subsection*{\textbf{Proof}}
To prove that the global model parameters \( w^t \) are bounded, we employ the energy function method. We construct an appropriate energy function and demonstrate that it does not increase with each iteration, thereby indirectly proving the boundedness of \( w^t \).

\subsubsection*{\textbf{Construction of the Energy Function}}
Define the energy function \( E^t \) as
\begin{equation}
E^t = \|w^t - w^*\|^2,
\end{equation}
where \( w^* \) is the optimal solution satisfying \( f(w^*) = f^* \).

\subsubsection*{\textbf{Evolution of the Energy Function}}
According to the model update rule, the global model parameters are updated as
\begin{equation}
w^{t+1} = w^t - \eta_g \Delta^t,
\end{equation}
where \( \eta_g \) is the global learning rate, and \( \Delta^t \) is the aggregated gradient update at step \( t \).
Therefore, the change in the energy function is
\begin{align}
E^{t+1} 
&= \|w^{t+1} - w^*\|^2 \nonumber\\
&= \|w^t - \eta_g \Delta^t - w^*\|^2 \nonumber\\
&= \|w^t - w^*\|^2 
  - 2\eta_g \langle w^t - w^*, \Delta^t \rangle 
  + \eta_g^2 \|\Delta^t\|^2.
\label{eq:energy_evolution}
\end{align}

\subsubsection*{\textbf{Bounding $\Delta^t$}}
Based on our update mechanism, $\Delta^t$ is an aggregation of updates from multiple clients:
\begin{equation}
\Delta^t = \frac{1}{C^t} \sum_{k \in \mathcal{S}_t} \Delta_k^t,
\end{equation}
where \( \mathcal{S}_t \) is the set of selected clients at step \( t \), and \( \Delta_k^t \) is the update from client \( k \) at step \( t \).
Each client's update \( \Delta_k^t \) can be expressed as
\begin{equation}
\Delta_k^t = \sum_{q=0}^{Q-1} \eta_\ell^{(q)} \, g_k\bigl(y_{k,q}^{t-\tau_k}\bigr),
\end{equation}
where \( \eta_\ell^{(q)} \) is the local learning rate for the \( q \)-th local step, and \( y_{k,q}^{t-\tau_k} \) is the local model parameter of client \( k \) at the \( q \)-th step with staleness \( \tau_k \).

\subsubsection*{\textbf{Expected Change in the Energy Function}}
Taking expectations, we obtain
\begin{align}
\mathbb{E}[E^{t+1}] 
&= \mathbb{E}[E^t]
   - 2\eta_g \,\mathbb{E}\bigl[\langle w^t - w^*, \Delta^t \rangle\bigr] 
   + \eta_g^2 \,\mathbb{E}\bigl[\|\Delta^t\|^2\bigr].
\label{eq:expected_energy_change}
\end{align}

\subsubsection*{\textbf{Analyzing \( \mathbb{E}\bigl[\langle w^t - w^*, \Delta^t \rangle\bigr] \)}}
Using Assumption \ref{assumption:unbiased} and the linearity of expectation,
\begin{align}
\mathbb{E}[\Delta^t] 
&= \frac{1}{C^t} \sum_{k \in \mathcal{S}_t} \mathbb{E}[\Delta_k^t] \nonumber\\
&= \frac{1}{C^t} \sum_{k \in \mathcal{S}_t} \sum_{q=0}^{Q-1} \eta_\ell^{(q)}\,\nabla F_k\bigl(y_{k,q}^{t-\tau_k}\bigr).
\label{eq:expected_delta}
\end{align}

Thus,
\begin{align}
\mathbb{E}\bigl[\langle w^t - w^*, \Delta^t \rangle\bigr] 
&= \frac{1}{C^t} \sum_{k \in \mathcal{S}_t} \sum_{q=0}^{Q-1} \eta_\ell^{(q)} 
   \langle w^t - w^*, \nabla F_k(y_{k,q}^{t-\tau_k}) \rangle.
\label{eq:inner_product}
\end{align}

Leveraging the \( L \)-smoothness of \( f \), we have
\begin{align}
f(w^{t+1}) &\leq f(w^t) 
    + \langle \nabla f(w^t), w^{t+1} - w^t \rangle 
    + \frac{L}{2} \|w^{t+1} - w^t\|^2.
\label{eq:f_smooth}
\end{align}

Substituting the update rule,
\begin{align}
f(w^{t+1}) &\leq f(w^t) 
    - \eta_g \langle \nabla f(w^t), \Delta^t \rangle 
    + \frac{L \,\eta_g^2}{2} \|\Delta^t\|^2.
\label{eq:f_update}
\end{align}

Taking expectations and rearranging,
\begin{align}
\mathbb{E}\bigl[f(w^{t+1})\bigr] 
&\leq \mathbb{E}\bigl[f(w^t)\bigr] 
    - \eta_g \,\mathbb{E}\bigl[\langle \nabla f(w^t), \Delta^t\rangle\bigr] \notag \\
    &\quad + \frac{L \,\eta_g^2}{2} \,\mathbb{E}\bigl[\|\Delta^t\|^2\bigr].
\label{eq:f_expectation}
\end{align}

Since \( f(w) \) is bounded below by \( f^* \), summing the above inequality from \( t = 0 \) to \( t = T-1 \) yields
\begin{align}
\sum_{t=0}^{T-1} \mathbb{E}\bigl[\langle \nabla f(w^t), \Delta^t\rangle\bigr]
&\leq \frac{f(w^0) - f^*}{\eta_g} 
  + \frac{L \,\eta_g}{2}\sum_{t=0}^{T-1}\mathbb{E}\bigl[\|\Delta^t\|^2\bigr].
\label{eq:cumulative_inequality}
\end{align}

\subsubsection*{\textbf{Bounding \( \mathbb{E}\bigl[\|\Delta^t\|^2\bigr] \)}}
Using Jensen's inequality and Assumptions \ref{assumption:variance} and \ref{assumption:bounded_gradient},
\begin{align}
\mathbb{E}\bigl[\|\Delta^t\|^2\bigr]
&= \mathbb{E}\Bigl[\Bigl\|\frac{1}{C^t}\sum_{k \in \mathcal{S}_t} \Delta_k^t\Bigr\|^2\Bigr] 
 \leq \frac{1}{(C^t)^2}\sum_{k \in \mathcal{S}_t} \mathbb{E}\bigl[\|\Delta_k^t\|^2\bigr].
\label{eq:jensen}
\end{align}

Furthermore,
\begin{align}
\mathbb{E}\bigl[\|\Delta_k^t\|^2\bigr] 
&= \mathbb{E}\Bigl[\Bigl\|\sum_{q=0}^{Q-1} \eta_\ell^{(q)}\,g_k\bigl(y_{k,q}^{t-\tau_k}\bigr)\Bigr\|^2\Bigr] \nonumber\\
&\leq \Bigl(\sum_{q=0}^{Q-1} (\eta_\ell^{(q)})^2\Bigr) \,
   \mathbb{E}\Bigl[\sum_{q=0}^{Q-1} \|g_k(y_{k,q}^{t-\tau_k})\|^2 \Bigr].
\label{eq:delta_k_bound}
\end{align}

According to Assumptions \eqref{assumption:variance} and \eqref{assumption:bounded_gradient},
\begin{equation}
\mathbb{E}\bigl[\|\Delta_k^t\|^2\bigr] \leq Q\,\beta(Q)\,\bigl(\sigma_\ell^2 + \sigma_g^2 + G\bigr),
\end{equation}
where \( \beta(Q) = \sum_{q=0}^{Q-1} (\eta_\ell^{(q)})^2 \).
Therefore,
\begin{equation}
\mathbb{E}\bigl[\|\Delta^t\|^2\bigr] 
\leq \frac{Q\,\beta(Q)}{(C^t)^2} \cdot C^t \cdot \bigl(\sigma_\ell^2 + \sigma_g^2 + G\bigr)
\leq \frac{Q\,\beta(Q)}{\mathcal{C}_{\text{min}}}\,\bigl(\sigma_\ell^2 + \sigma_g^2 + G\bigr).
\end{equation}

\subsubsection*{\textbf{Combining the Results}}
Substituting the bound on \( \mathbb{E}\bigl[\|\Delta^t\|^2\bigr] \) into \eqref{eq:cumulative_inequality}, we obtain
\begin{align}
\sum_{t=0}^{T-1} \mathbb{E}\bigl[\langle \nabla f(w^t), \Delta^t\rangle\bigr]
&\leq \frac{f(w^0) - f^*}{\eta_g} 
  + \frac{L \,\eta_g\,Q\,\beta(Q)}{2\,\mathcal{C}_{\text{min}}}\,T \,\bigl(\sigma_\ell^2 + \sigma_g^2 + G\bigr).
\label{eq:combining_results_1}
\end{align}

On the other hand, applying the Cauchy-Schwarz inequality,
\begin{align}
\langle \nabla f(w^t), \Delta^t \rangle 
&\geq \frac{1}{2}\|\nabla f(w^t)\|^2 - \frac{1}{2}\|\Delta^t\|^2.
\label{eq:cs_inequality}
\end{align}

Taking expectations,
\begin{align}
\mathbb{E}\bigl[\langle \nabla f(w^t), \Delta^t \rangle\bigr] 
&\geq \frac{1}{2}\,\mathbb{E}\bigl[\|\nabla f(w^t)\|^2\bigr]
   - \tfrac{1}{2}\,\mathbb{E}\bigl[\|\Delta^t\|^2\bigr].
\label{eq:cs_inequality_exp}
\end{align}

Therefore,
\begin{align}
\sum_{t=0}^{T-1} \mathbb{E}\bigl[\|\nabla f(w^t)\|^2\bigr]
&\leq 2 \sum_{t=0}^{T-1} \mathbb{E}\bigl[\langle \nabla f(w^t), \Delta^t \rangle\bigr]
   + \sum_{t=0}^{T-1} \mathbb{E}\bigl[\|\Delta^t\|^2\bigr].
\label{eq:nabla_f_bound}
\end{align}

Combining \eqref{eq:combining_results_1} and \eqref{eq:nabla_f_bound},
\begin{align}
\sum_{t=0}^{T-1} \mathbb{E}\bigl[\|\nabla f(w^t)\|^2\bigr]
&\leq 
2 \Bigl( 
   \frac{f(w^0) - f^*}{\eta_g} 
   + \frac{L \,\eta_g \, Q \,\beta(Q)}{2\,\mathcal{C}_{\text{min}}}\,T\,(\sigma_\ell^2 + \sigma_g^2 + G)
\Bigr)\nonumber\\
&\quad+ \frac{Q\,\beta(Q)}{\mathcal{C}_{\text{min}}}\,T\,(\sigma_\ell^2 + \sigma_g^2 + G).
\label{eq:final_bound}
\end{align}

Simplifying,
\begin{align}
\frac{1}{T}\sum_{t=0}^{T-1} \mathbb{E}\bigl[\|\nabla f(w^t)\|^2\bigr]
\leq \frac{2\bigl(f(w^0) - f^*\bigr)}{\eta_g\,T} + \left(\frac{L \,\eta_g\,Q\,\beta(Q)}{\mathcal{C}_{\text{min}}} + \frac{Q\,\beta(Q)}{\mathcal{C}_{\text{min}}}\right)\,(\sigma_\ell^2 + \sigma_g^2 + G).
\label{eq:final_bound_simplified}
\end{align}

\section*{Proof of the Rank of a Gaussian Noise Matrix}

\subsection*{\textbf{Viewing Matrices as a Collection of Row Vectors}}

Let $M$ and $N$ be both $n \times n$ real matrices. Define
\begin{equation}
M =
\begin{pmatrix}
\mathbf{m}_1 \\
\mathbf{m}_2 \\
\vdots \\
\mathbf{m}_n
\end{pmatrix}, \quad
N =
\begin{pmatrix}
\mathbf{n}_1 \\
\mathbf{n}_2 \\
\vdots \\
\mathbf{n}_n
\end{pmatrix}.
\end{equation}
where each $\mathbf{m}_i$ and $\mathbf{n}_i$ is a row vector (in $\mathbb{R}^n$).

Then
\begin{equation}
A = M + N =
\begin{pmatrix}
\mathbf{m}_1 + \mathbf{n}_1 \\
\mathbf{m}_2 + \mathbf{n}_2 \\
\vdots \\
\mathbf{m}_n + \mathbf{n}_n
\end{pmatrix}.
\end{equation}

We want to investigate whether $A$ has full rank (i.e., $\operatorname{rank}(A) = n$).

\subsection*{\textbf{Step-by-Step Determination of Linear Independence}}

For $A$ to have full rank, it is equivalent to ensuring that the $n$ row vectors $\{\mathbf{m}_1 + \mathbf{n}_1, \mathbf{m}_2 + \mathbf{n}_2, \dots, \mathbf{m}_n + \mathbf{n}_n\}$ are linearly independent (i.e., they span $\mathbb{R}^n$).

\begin{itemize}
    \item Is the first row vector $\mathbf{m}_1 + \mathbf{n}_1$ equal to 0?
    
    Since $\mathbf{n}_1$ is a continuous random vector, the probability that it falls on a specific point (such as exactly canceling out $\mathbf{m}_1$ to make a zero vector) is 0. Thus, the probability of $\mathbf{m}_1 + \mathbf{n}_1 = 0$ is 0.

    In other words, with probability 1, we obtain a nonzero vector as the first row.

    \item Suppose we already know that the first $k-1$ rows $(1 \leq k - 1 < n)$ are linearly independent. Now, consider the $k$-th row. To maintain linear independence with the previous $k-1$ rows, we require that
    \begin{equation}
    \mathbf{m}_k + \mathbf{n}_k \notin \operatorname{span} \{\mathbf{m}_1 + \mathbf{n}_1, \mathbf{m}_2 + \mathbf{n}_2, \dots, \mathbf{m}_{k-1} + \mathbf{n}_{k-1} \}.
    \end{equation}
    
    In $\mathbb{R}^n$, $\operatorname{span}\{\dots\}$ is a linear subspace of dimension at most $k-1$ (at most $k-1$-dimensional, while $n \geq k$). A continuous random vector (i.e., the value of $\mathbf{n}_k$) has a probability of 0 of falling exactly into a lower-dimensional subspace.

    \item Therefore, with probability 1, the $k$-th row will not fall into the subspace spanned by the previous $k-1$ rows, thereby preserving (and expanding) linear independence.
\end{itemize}

\subsection*{\textbf{Recursive Conclusion: Probability of Full Rank is 1}}

From the above step-by-step analysis, we conclude that under the assumption that the noise vectors $\mathbf{n}_i$ are continuous random vectors, each row vector has probability 1 of maintaining linear independence with the preceding row vectors. Therefore, in the end, all row vectors will be linearly independent, meaning the matrix reaches full rank with probability 1. That is:
\begin{equation}
\Pr \left[ \operatorname{rank}(M + N) = n \right] = 1.
\end{equation}

\bibliographystyle{IEEEtran} 
\bibliography{citation}

\end{document}